\documentclass{article} % For LaTeX2e
\usepackage[dvipsnames]{xcolor}
\usepackage{iclr2021_conference,times}

\usepackage{placeins}
\usepackage{hyperref}
\usepackage{url}
\usepackage{epsfig}
\usepackage{graphicx}
\usepackage{amsmath}
\usepackage{amssymb}
\usepackage{tikz}
\usepackage{amsthm}
\usepackage{algorithm,algorithmic} % @max added for algorithm support
\usepackage{algorithmic,eqparbox,array}% @max for aligned algorithm comments
\usepackage{tikz} % To generate the plot from csv
\usepackage{pgfplots}
\usepackage{layouts}

\title{A straightforward line search approach on the expected empirical loss for stochastic deep learning problems}

\author{Maximus Mutschler \& Andreas Zell, \\
Department of Computer Science\\
University of T\"ubingen \\
Sand 1, D-72076 T\"ubingen, Germany \\
\texttt{\{maximus.mutschler, andreas.zell\}@uni-tuebingen.de} \\
}

\iclrfinalcopy % Uncomment for camera-ready version, but NOT for submission.
\begin{document}

\maketitle
\vspace{-0.8cm}
\begin{abstract}
	\vspace{-0.3cm}
   A fundamental challenge in deep learning is that the optimal step sizes for update steps of stochastic gradient descent are unknown. In traditional optimization, line searches are used to determine good step sizes, however, in deep learning, it is too costly to search for good step sizes on the expected empirical loss due to noisy losses. This empirical work shows that it is possible to approximate the expected empirical loss on vertical cross sections for common deep learning tasks considerably cheaply. This is achieved by applying traditional one-dimensional function fitting to measured noisy losses of such cross sections. The step to a minimum of the resulting approximation is then used as step size for the optimization. This approach leads to a robust and straightforward optimization method which performs well across datasets and architectures without the need of hyperparameter tuning.
\end{abstract}
\vspace{-0.5cm}
\section{Introduction and Background}
\vspace{-0.3cm}
The automatic determination of an optimal learning rate schedule to train models with stochastic gradient descent or similar optimizers is still not solved satisfactorily for standard and especially new deep learning tasks. Frequently, optimization approaches utilize the information of the loss and gradient of a single batch to perform an update step. However, those approaches focus on the batch loss, whereas the optimal step size should actually be determined for the  empirical loss, which is the expected loss over all batches. In classical optimization line searches are commonly used to determine good step sizes.  In deep learning, however, the noisy loss functions makes it impractically costly to search for step sizes on the  empirical loss. This work empirically revisits that the  empirical loss has a simple shape in the direction of noisy gradients. Based on this information, it is shown that the  empirical loss can be easily fitted with lower order polynomials in these directions. This is done by performing a straightforward, one-dimensional regression on batch losses sampled in such a direction. It then becomes simple to determine a suitable minimum and thus a good step size from the approximated function. This results in  a line search on the  empirical loss. Compared to the direct measurement of the  empirical loss on several locations, our approach is cost-efficient since it solely requires a sample size of about 500 losses to approximate a cross section of the  loss. From a practical point of view this is still too expensive to determine the step size for each step. Fortunately, it turns out to be sufficient to estimate a new step size only a few times during a training process, which, does not require any additional time due to more beneficial update steps. We show that this straightforward optimization approach called ELF (Empirical Loss Fitting optimizer), performs robustly across datasets and models without the need for hyperparameter tuning. This makes ELF a choice to be considered in order to achieve good results for new deep learning tasks out of the box.

In the following we will revisit the fundamentals of optimization in deep learning to make our approach easily understandable. Following \cite{goodfellow2016deep}, the aim of optimization in deep learning generally means to find a global minimum of the true loss (risk) function $\mathcal{L}_{true}$ which is the expected loss over all elements of the data generating distribution $p_{data}$:
\begin{equation}
\mathcal{L}_{true}(\mathbf{\theta})=\mathbb{E}_{(\mathbf{x},y)\sim p_{data}}L(f(\mathbf{x};\mathbf{\theta}),y)
\end{equation}
where $L$ is the loss function for each sample $(\mathbf{x},y)$, $\mathbf{\theta}$ are the parameters to optimize and $f$ the model function. However, $p_{data}$ is usually unknown and we need to use an empirical approximation $\hat{p}_{data}$, which is usually indirectly given by a dataset $\mathbb{T}$. Due to the central limit theorem we can assume $\hat{p}_{data}$ to be Gaussian. In practice optimization is performed on the empirical loss  $\mathcal{L}_{emp}$:
\begin{align}
\mathcal{L}_{emp}(\mathbf{\theta})&=\mathbb{E}_{(\mathbf{x},y)\sim \hat{p}_{data}}L(f(\mathbf{x};\mathbf{\theta}),y)=\frac{1}{|\mathbb{T}|}\sum\limits_{(\mathbf{x},y)\in \mathbb{T}}L(f(\mathbf{x};\mathbf{\theta}),y)
\end{align}
An unsolved task is to find a global minimum of $\mathcal{L}_{true}$ by optimizing on $\mathcal{L}_{emp}$ if $|\mathbb{T}|$ is finite. Thus, we have to assume that a small value of $\mathcal{L}_{emp}$ will also be small for $\mathcal{L}_{true}$.
Estimating $\mathcal{L}_{emp}$ is impractical and expensive, therefore we approximate it with mini batches:
\begin{align}
\mathcal{L}_{batch}(\mathbf{\theta},\mathbb{B})&=\frac{1}{|\mathbb{B}|}\sum\limits_{(\mathbf{x},y)\in \mathbb{B} \subset \mathbb{T}}L(f(\mathbf{x};\mathbf{\theta}),y)
%\nabla_\mathbf{\theta}\mathcal{L}_{batch}(\mathbf{\theta},\mathbb{B})&=\frac{1}{|\mathbb{B}|}\sum\limits_{(\mathbf{x},y)\in \mathbb{B} \subset T}\nabla_\mathbf{\theta}L(f(\mathbf{x};\mathbf{\theta}),y)
\end{align} where $\mathbb{B}$ denotes a batch.  We call the dataset split in batches $\mathbb{T}_{batch}$.\\
We now can reinterpret $\mathcal{L}_{emp}$ as the empirical mean value over a list of losses $\mathbb{L}$ which includes the output of $\mathcal{L}_{batch}(\mathbf{\theta},\mathbb{B})$ for each batch $\mathbb{B}$:
\begin{align}
\mathcal{L}_{emp}(\mathbf{\theta})&=\frac{1}{|\mathbb{L}|}\sum\limits_{\mathcal{L}_{batch}(\mathbf{\theta},\mathbb{B})\in \mathbb{L}}\mathcal{L}_{batch}(\mathbf{\theta},\mathbb{B})
\end{align}
%\subsection{Line Searches}
 A vertical cross section $l_{emp}(s)$ of $\mathcal{L}_{emp}(\mathbf{\theta})$ in the direction $d$ through the parameter vector $\mathbf{\theta}_0$ is given by
\begin{align}
l_{emp}(s;\mathbf{\theta}_0,d)=\mathcal{L}_{emp}(\mathbf{\theta}_0+s \cdot \mathbf{d})
\end{align}
For simplification, we refer to $l$ as line function or cross section.
The step size to the minimum of $l_{emp}(s)$ is called $s_{min}$.

Many direct and indirect line search approaches for deep learning are often applied on $\mathcal{L}_{batch}(\mathbf{\theta},\mathbb{B})$ (\cite{pal}, \cite{L4_alternative}, \cite{L4}, \cite{hypergradientdescent},\cite{backtracking_line_search_NIPS}). \quad
\cite{pal} approximate an exact line search, which implies estimating the global minimum of a line function, by using one-dimensional parabolic approximations. The other approaches, directly or indirectly, perform inexact line searches by estimating positions of the line function, which fulfill specific conditions, such as the Goldberg, Armijo and Wolfe conditions  (\cite{numerical_optimization}).
However, \cite{pal} empirically suggests that line searches on $\mathcal{L}_{batch}$ are not optimal since minima of line functions of $\mathcal{L}_{batch}$  are not always good estimators for the minima of line functions of $\mathcal{L}_{emp}$. Thus, it seems more promising to perform a line search on $\mathcal{L}_{emp}$.
This is cost intensive since we need to determine $L(f(\mathbf{x};\mathbf{\theta}_0+s\cdot \mathbf{d} ),y)$ for all $(\mathbf{x},y)\in \mathbb{T}$ for multiple s of a line function. Probabilistic Line Search (PLS) (\cite{probabilisticLineSearch}) addresses this problem by performing Gaussian process regressions, which result in multiple one dimensional cubic splines. In addition, a probabilistic belief over the first  (= Armijo condition) and second Wolfe condition is introduced to find good update positions. The major drawback of this conceptually appealing but complex method is, that for each batch the squared gradients of each input sample have to be computed. This is not supported by default by common deep learning libraries and therefore has to be  implemented manually for every layer in the model, which makes its application impractical. Gradient-only line search (GOLS1) (\cite{gradientOnlyLineSearch}) pointed out empirically that the noise of directional derivatives in negative gradient direction is considerably smaller than the noise of the losses. They argue that they can approximate a line search on $\mathcal{L}_{emp}$ by considering consecutive noisy directional derivatives.
Adaptive methods, such as \cite{adam} \cite{AdaBound} \cite{AmsGrad} \cite{radam} \cite{rmsProp} \cite{adadelta} \cite{grad_descent} concentrate more on finding good directions than on optimal step sizes. Thus, they could benefit from line search approaches applied on their estimated directions.
Second order methods, such as \cite{S-LBFGS} \cite{oLBFGS} \cite{KFAC} \cite{L-sr1} \cite{gausnewton} tend to find better directions but are generally too expensive for deep learning scenarios.

Our approach follows PLS and GOLS1 by performing a line search directly on $\mathcal{L}_{emp}$. We use a regression on multiple $\mathcal{L}_{batch}(\mathbf{\theta}_0+s\cdot \mathbf{d},\mathbb{B})$ values sampled with different step sizes $s$ and different batches $\mathbb{B}$, to estimate a minimum of a line function of $\mathcal{L}_{emp}$  in direction $\mathbf{d}$. Consequently, this work is a further step towards efficient steepest descent line searches on $\mathcal{L}_{emp}$, which show linear convergence on any deterministic function that is twice continuously differentiable, has a relative minimum and only positive eigenvalues of the Hessian at the minimum (see \cite{nonlinear_programming}). 
The details as well as the empirical foundation of our approach are explained in the following.

% PAL  consider the optimal scenario, an tries to find the exact minimum on a line with the help of parabolic approximations, whereas others use inexact line searches by using various conditions (Goldberg, Armijo, Wolfe \cite{numerical_optimization}) to find good enough steps.
%Considering the scenario where we want to find the exact minimum on a line, the empirical results of  Todo Cite PAL suggest that line searches like on $\mathcal{L}_{batch}(\theta,B)$ are not optimal since minima of ${L}_{btach}$ on lines are not always a good estimator for the minima ${L}_{emp}$. Thus, the aim is, to perform a line search on $\mathcal{L}_{emp}$. This is cost intensive since we need to determine every $L(f(\mathbf{x};\mathbf{\theta}_0+s\cdot d ),y)$ for all $(\mathbf{x},y)\in T$ for multiple s on a line. PLS tackled this problem by performing a Gaussian process regression which results in multiple one dimensional cubic splines. In addition, a probabilistic believe over the first  (= Armijo) and second Wolfe condition is introduced to find good update positions.(Todo read this again). The major drawback of this beautiful and complex method is, that for each batch the squared gradients of each input sample have to be computed, which is per default not supported by standard deep learning libraries and has to be manually implemented for every layer in the model, which makes it impractical to use. 

\section{Our approach}
\subsection{empirical foundations}
\label{subsec:empirical_foundations}
\begin{figure*}[t!]
	\centering
	\newcommand\disfigwidth{0.3}
	\vspace{-0.3cm}
	\includegraphics[width=\disfigwidth\linewidth]{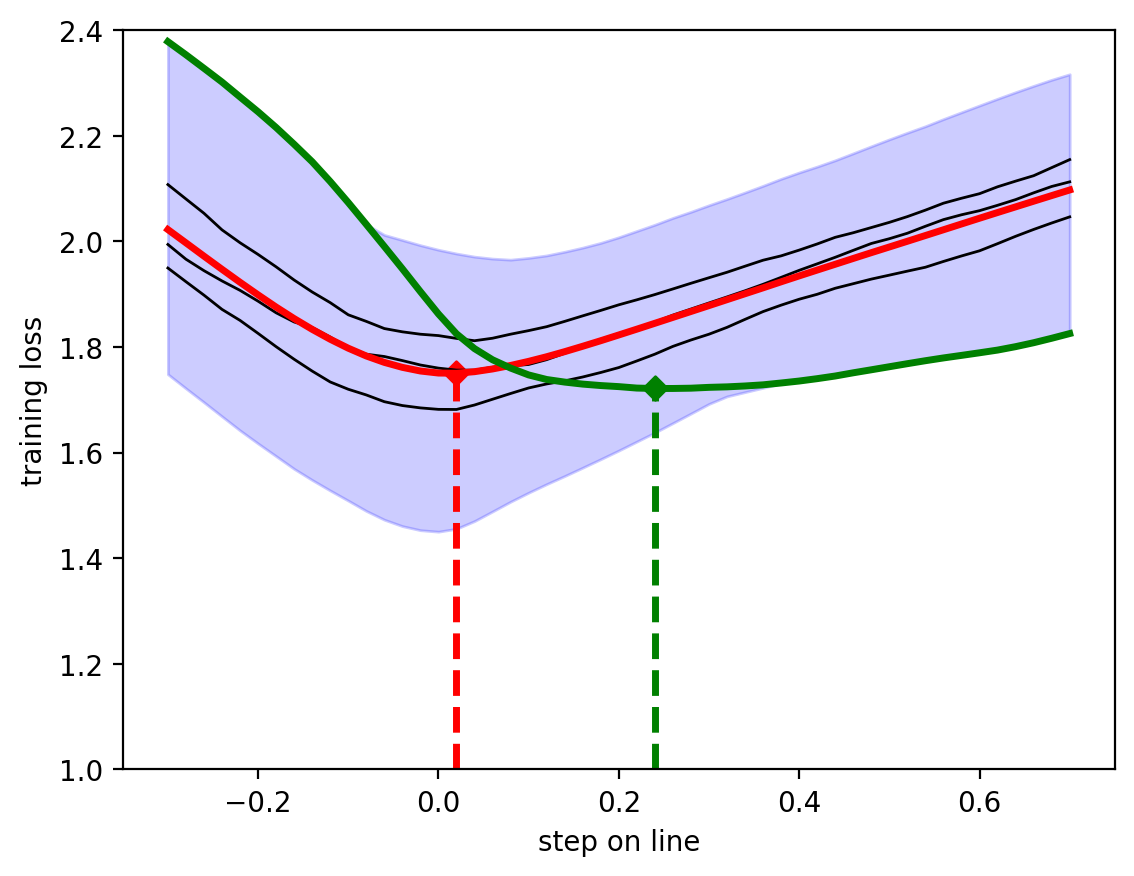}
	\includegraphics[width=\disfigwidth\linewidth]{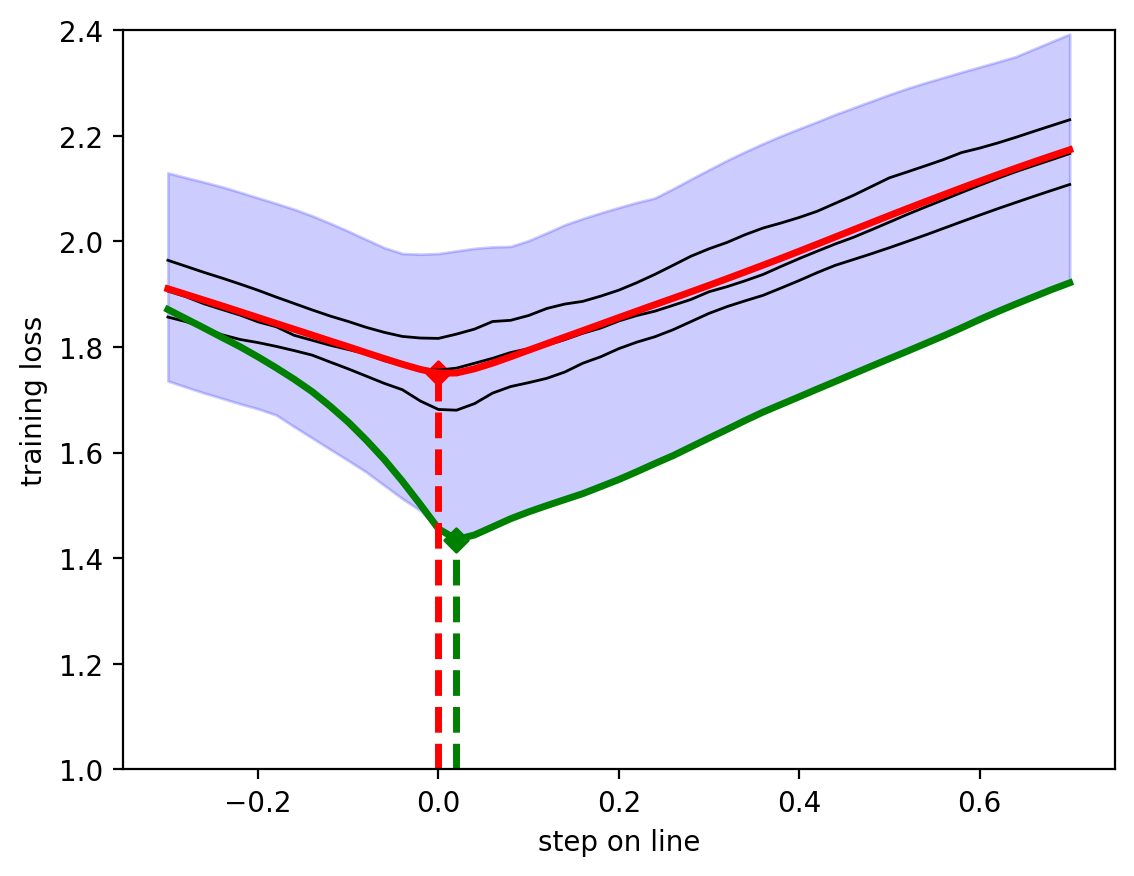}
	\includegraphics[width=\disfigwidth\linewidth]{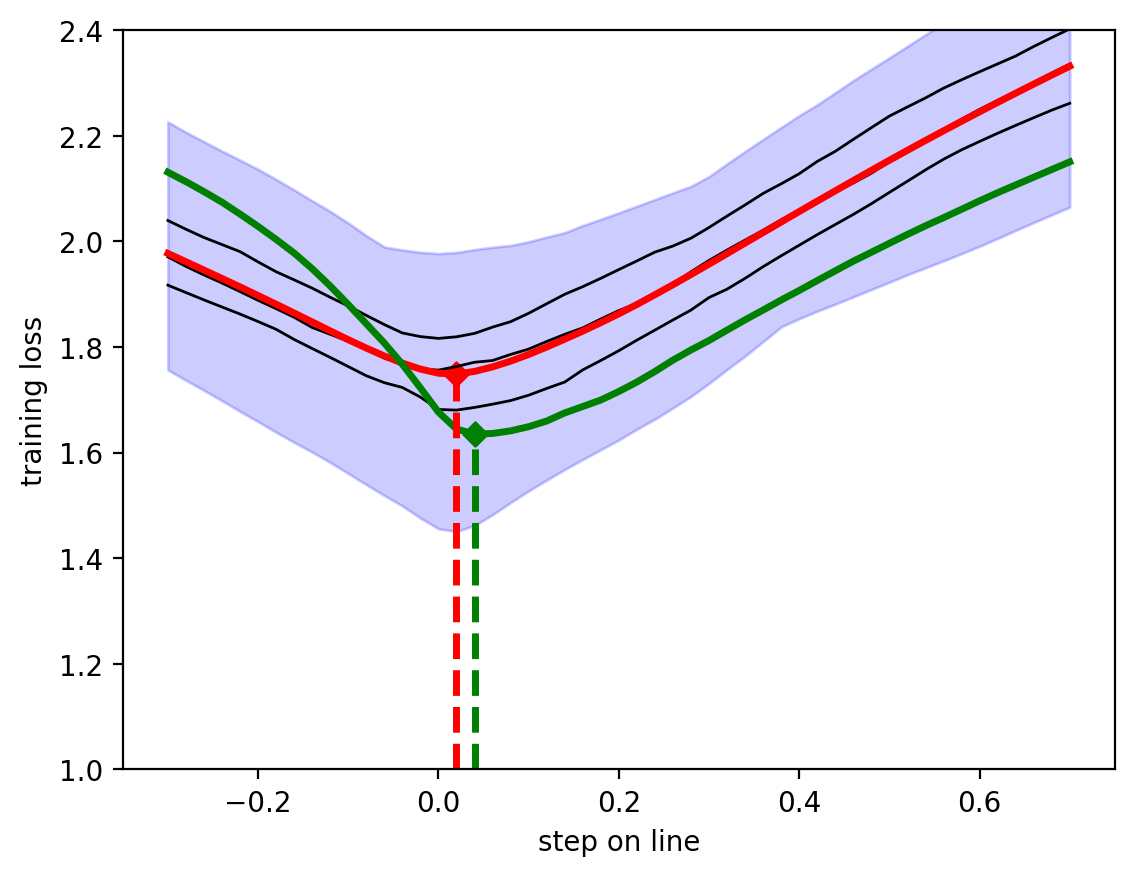}\\
	\vspace{-0.1cm}
	\includegraphics[width=\disfigwidth\linewidth]{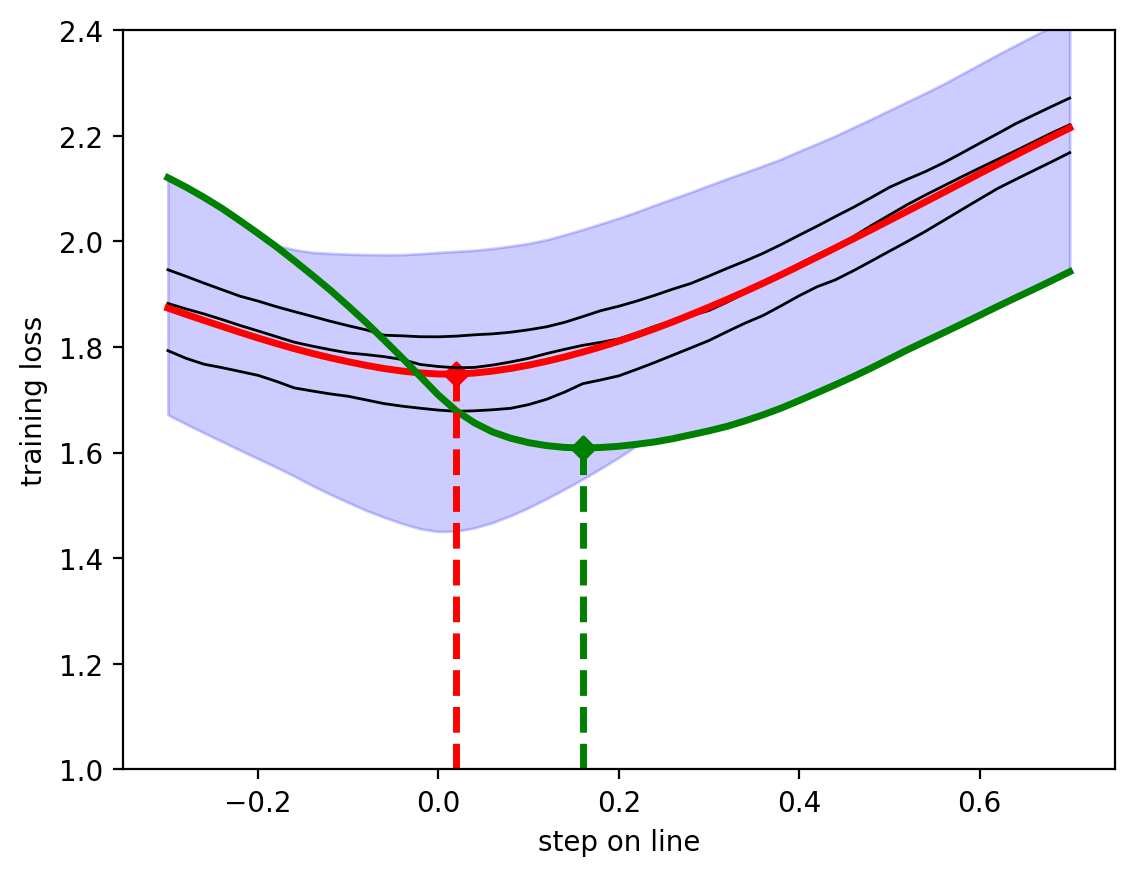}
	\includegraphics[width=\disfigwidth\linewidth]{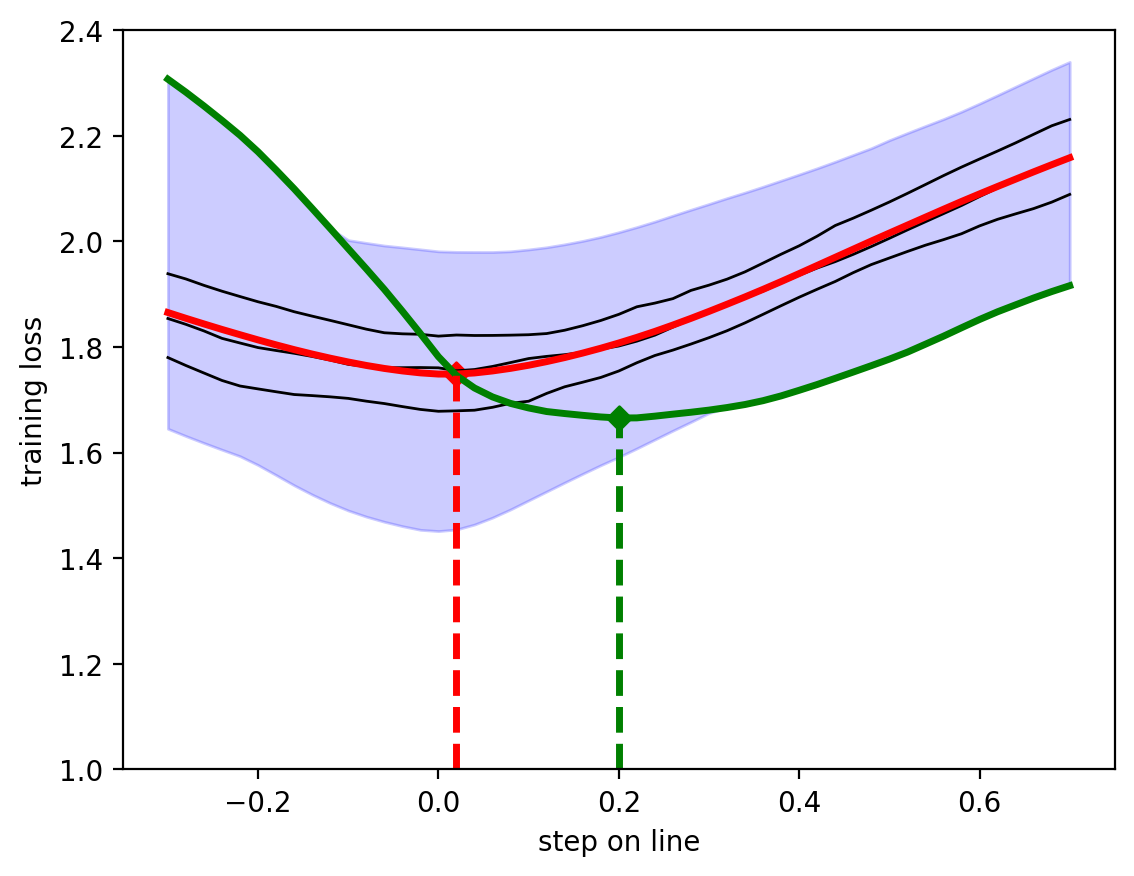}
	\includegraphics[width=\disfigwidth\linewidth]{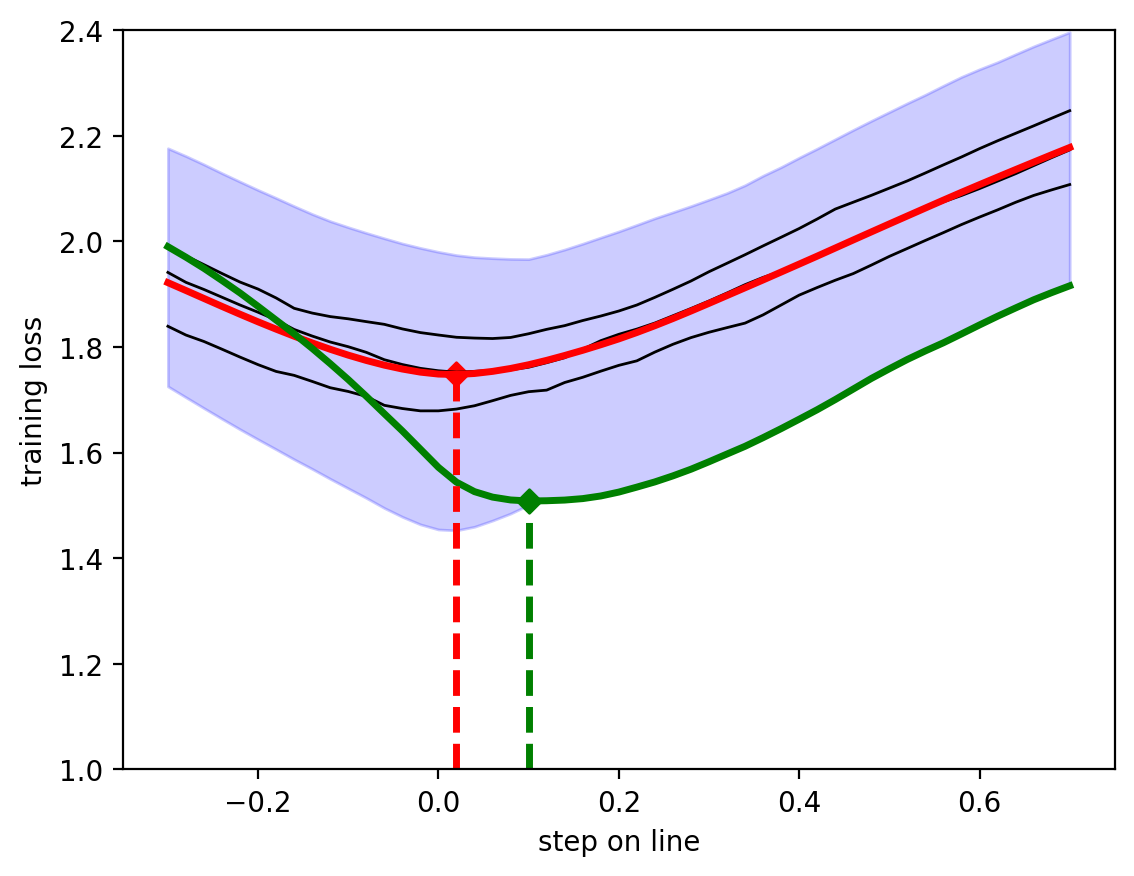}\\
	\vspace{-0.25cm}
	%\vspace{-0.2cm}
	%\includegraphics[width=\disfigwidth\linewidth]{plot_data/distribution_plots/train_line306}
	%\includegraphics[width=\disfigwidth\linewidth]{plot_data/distribution_plots/train_line307}
	%\includegraphics[width=\disfigwidth\linewidth]{plot_data/distribution_plots/train_line308}
\caption{Distributions over all batch losses $\mathcal{L}_{batch}$ (\textcolor{BlueViolet}{blue}) on consecutive and representative cross sections during a training process of a ResNet32 on CIFAR-10. The empirical loss $\mathcal{L}_{emp}$, is given in \textcolor{red}{red}, the quartiles in black. The batch loss, whose negative gradient defines the search direction, is given in \textcolor{ForestGreen}{green}. See Section \ref{subsec:empirical_foundations} for interpretations.\vspace{-0.25cm}}
\label{fig:distributionplots}
\end{figure*}

\cite{walkwithsgd,pal,probabilisticLineSearch,empericalLineSearchApproximations} showed empirically that line functions of $\mathcal{L}_{batch}$ in negative gradient directions tend to exhibit a simple shape for all analyzed deep learning problems.
To get an intuition of how lines of the empirical loss in the direction of the negative gradient tend to behave, we tediously sampled $\mathcal{L}_{batch}(\mathbf{\theta}_t+s\cdot - \nabla_{\mathbf{\theta}_t}\mathcal{L}_{batch}(\mathbf{\theta}_t,\mathbb{B}_t) ),\mathbb{B})$ for 50 equally distributed $s$ between $-0.3$ and $0.7$ and every $\mathbb{B} \in \mathbb{T}$ for a training process of a ResNet32 trained on CIFAR-10 with a batch size of 100. The results are given in Figure \ref{fig:distributionplots}. 
\footnote{\scriptsize These results have already been published by the authors
 of this paper in another context in \cite{pal}} %\cite{pal}}
\\The results lead to the following characteristics: \textbf{1.} $l_{emp}$ has a simple shape and can be approximated well by lower order polynomials, splines or fourier series. \textbf{2.} $l_{emp}$  does not change much over consecutive lines. \textbf{3.} Minima of lines of $\mathcal{L}_{batch}$  can be shifted from the minima of $l_{emp}$ lines and can even lead to update steps which increase $\mathcal{L}_{emp}$.
Characteristic 3 consolidates why line searches on $\mathcal{L}_{emp}$ are to be favored over line searches on $\mathcal{L}_{batch}$. Although we derived these findings only from one training process, we can assure, by analyzing the measured point clouds of our approach, that they seem to be valid for all datasets, tasks, and models considered (see Appendix \ref{app:furtherLineApp}).

\subsection{OUR LINE SEARCH ON THE EXPECTED EMPIRICAL LOSS }
There are two major challenges to be solved in order to perform line searches on $\mathcal{L}_{emp}$:
\begin{enumerate}
	\item To measure $l_{emp}(s;\mathbf{\theta},\mathbf{d})$ it is required to determine every $L(f(\mathbf{x};\mathbf{\theta}_0+s\cdot \mathbf{d}),y)$ for all $(\mathbf{x},y)\in \mathbb{T}$ for all steps sized $s$ on a line.
	\item For a good direction of the line function one has to know\\ $\nabla_\mathbf{\theta}\mathcal{L}_{emp}(\mathbf{\theta})=\frac{1}{|\mathbb{T}|}\sum\limits_{\mathbb{B} \in \mathbb{T}}\nabla\mathcal{L}_{batch}(\mathbf{\theta},\mathbb{B})$. 
\end{enumerate}
We solve the first challenge by fitting $l_{emp}$ with lower order polynomials, which can be achieved accurately by sampling a considerably low number of batch loss values. 
We do not have an efficient solution for the second challenge, thus we have to simplify the problem by taking the unit gradient of the current batch $\mathbb{B}_t$, which is $\hat{\nabla}_\mathbf{\theta}\mathcal{L}_{batch}(\mathbf{\theta},\mathbb{B}_t)$, as the direction of the line search.
The line  function we search on is thus given by:
\begin{align}
l_{ELF}(s;\mathbf{\theta}_t,\mathbb{B}_t)=\mathcal{L}_{emp}(\mathbf\mathbf{{\theta}_t}+s\cdot-\hat{\nabla}_{\mathbf{\theta}_t}\mathcal{L}_{batch}({\mathbf{\theta}_t},\mathbb{B}_t))\approx \text{lower order polynomial}
\end{align}Note that  $\mathbf{\theta}_t,\mathbb{B}_t$ are fixed during the line search.

\begin{figure}[t!]
	\centering
	\newcommand\disfigwidth{0.3}
	\vspace{-0.3cm}
	\includegraphics[width=\disfigwidth\linewidth]{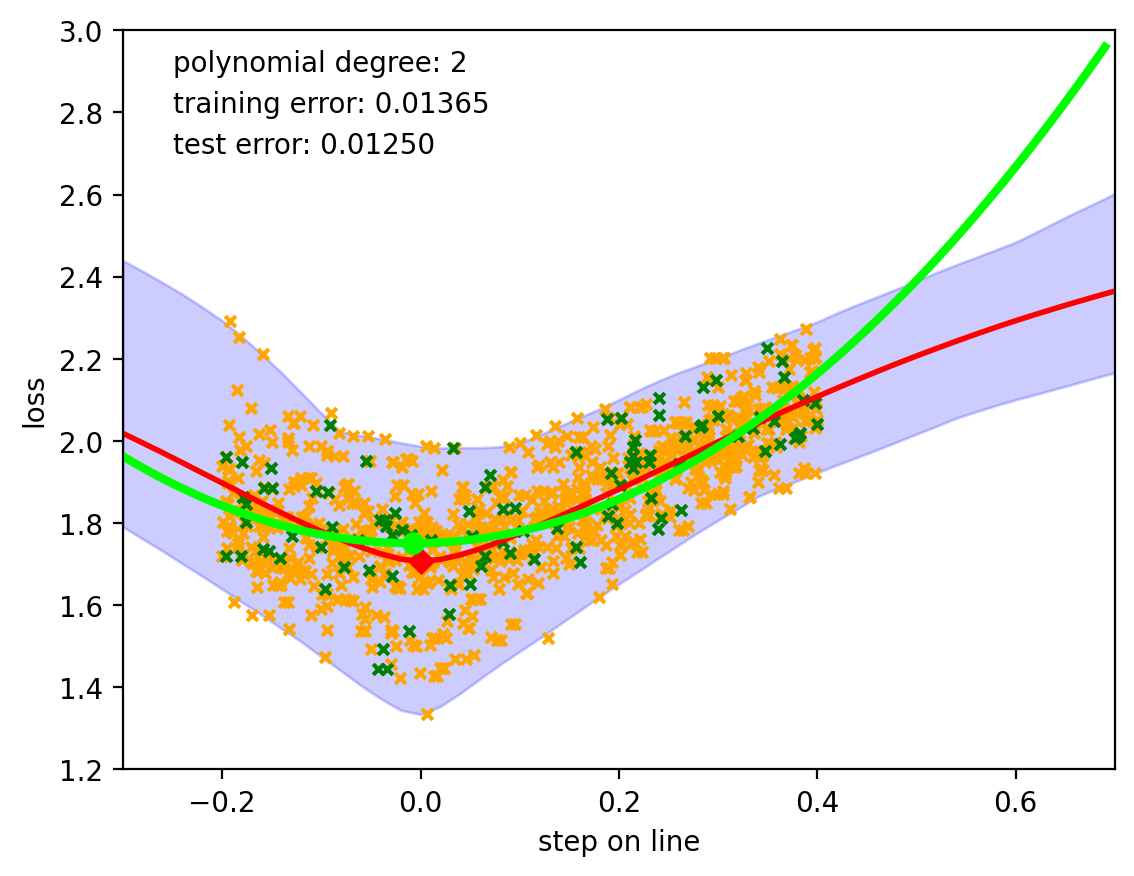}
	\includegraphics[width=\disfigwidth\linewidth]{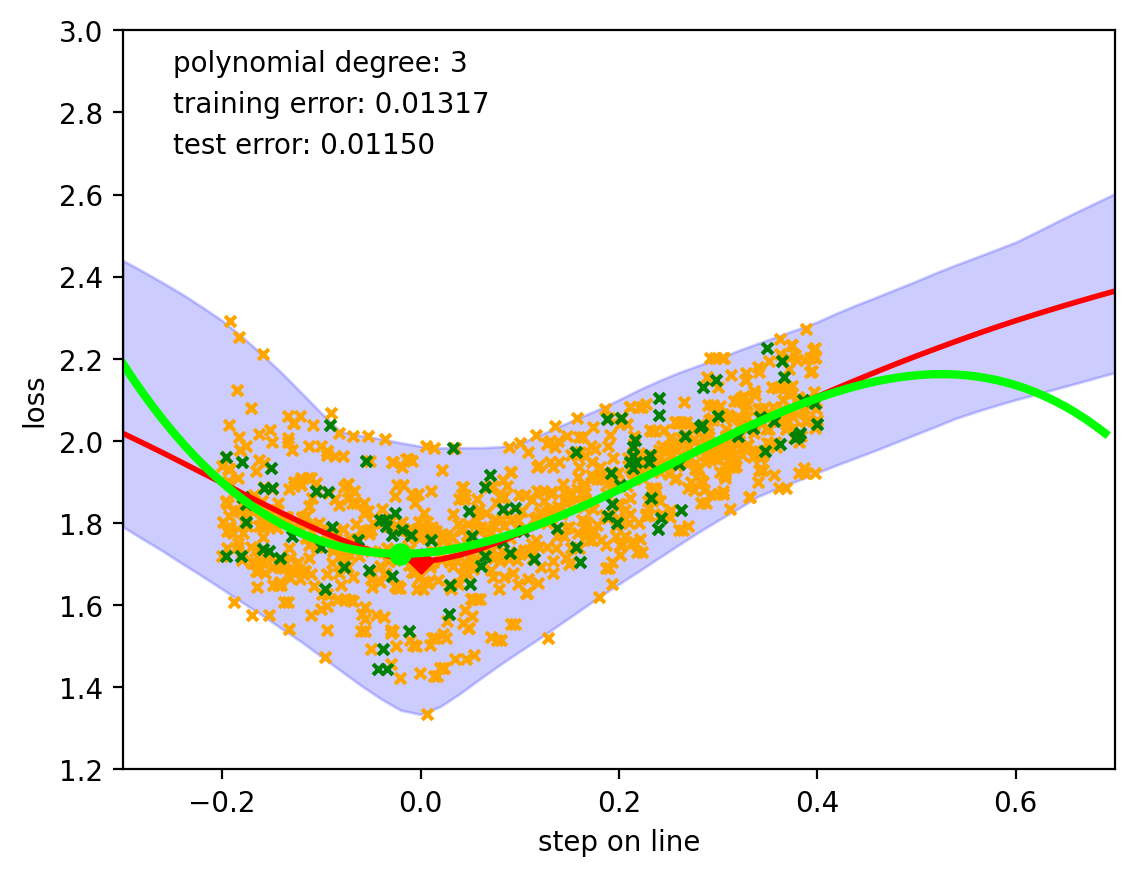}
	\includegraphics[width=\disfigwidth\linewidth]{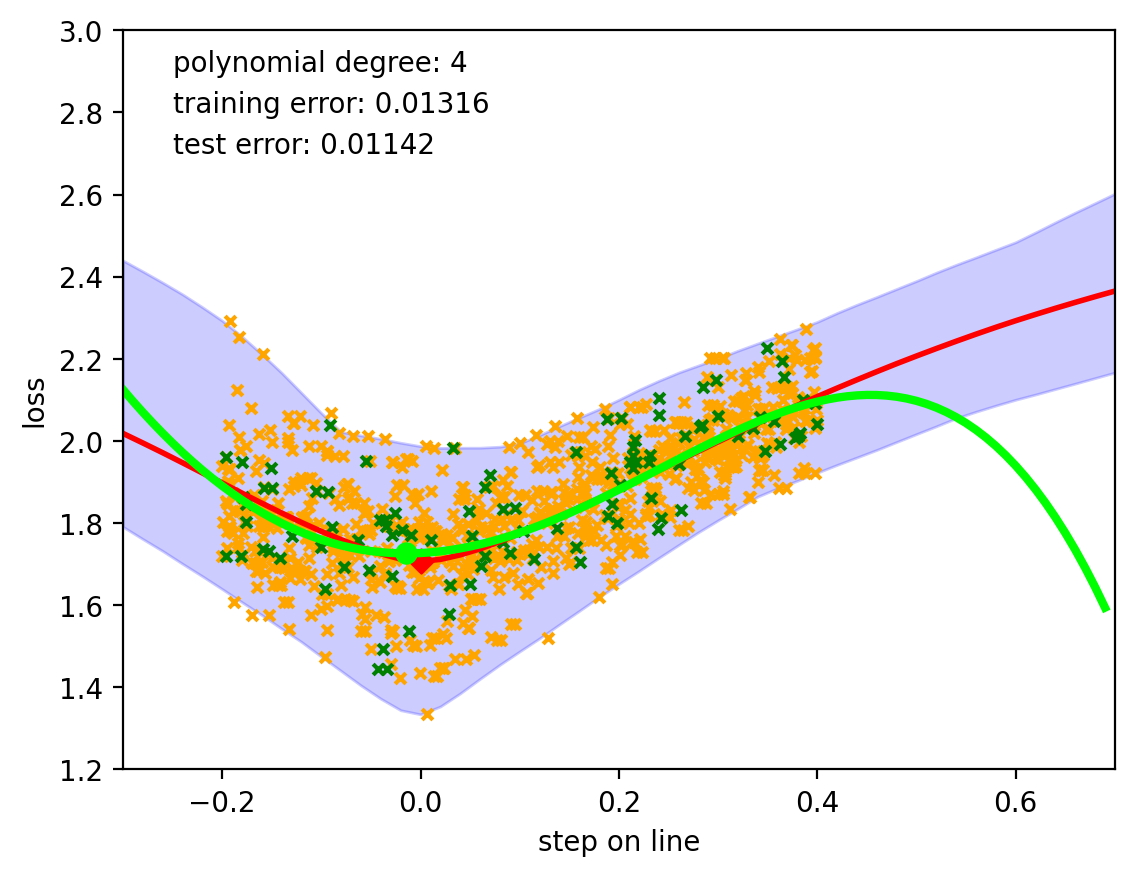}\\
	\vspace{-0.1cm}
	\includegraphics[width=\disfigwidth\linewidth]{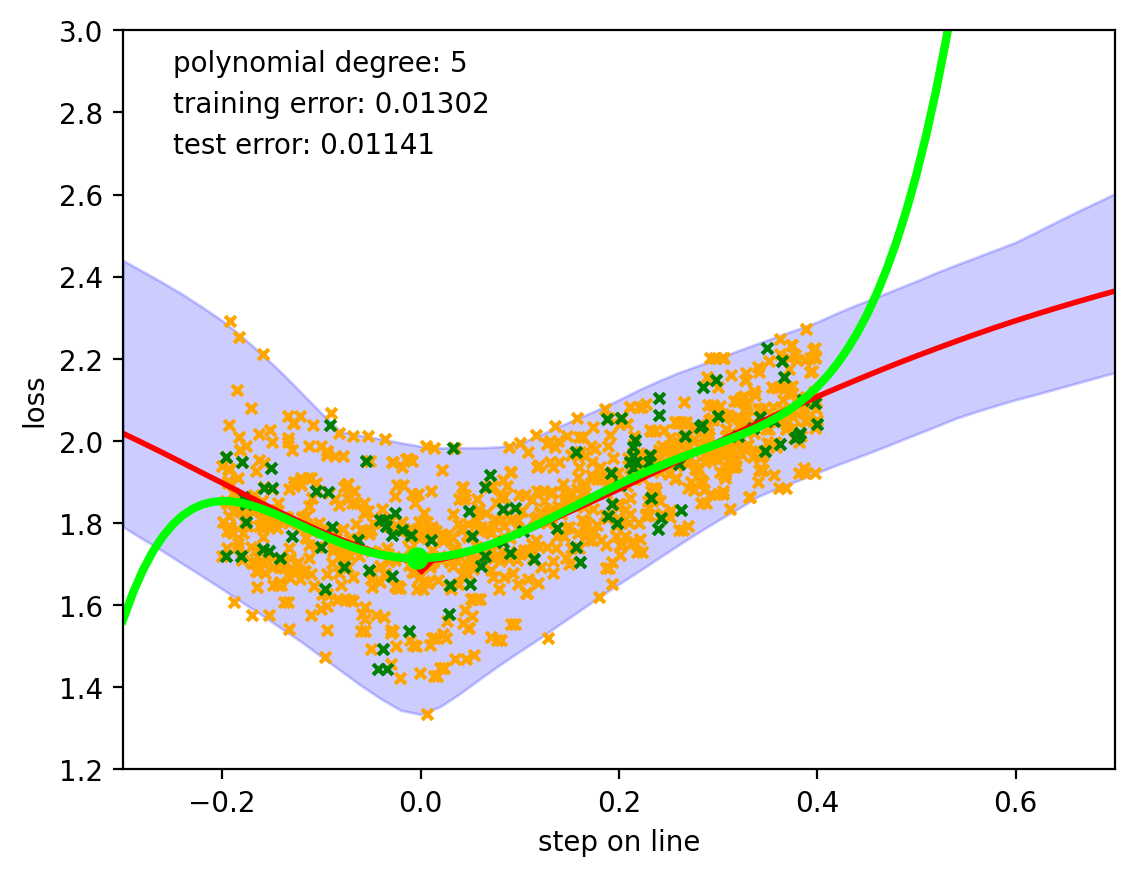}
	\includegraphics[width=\disfigwidth\linewidth]{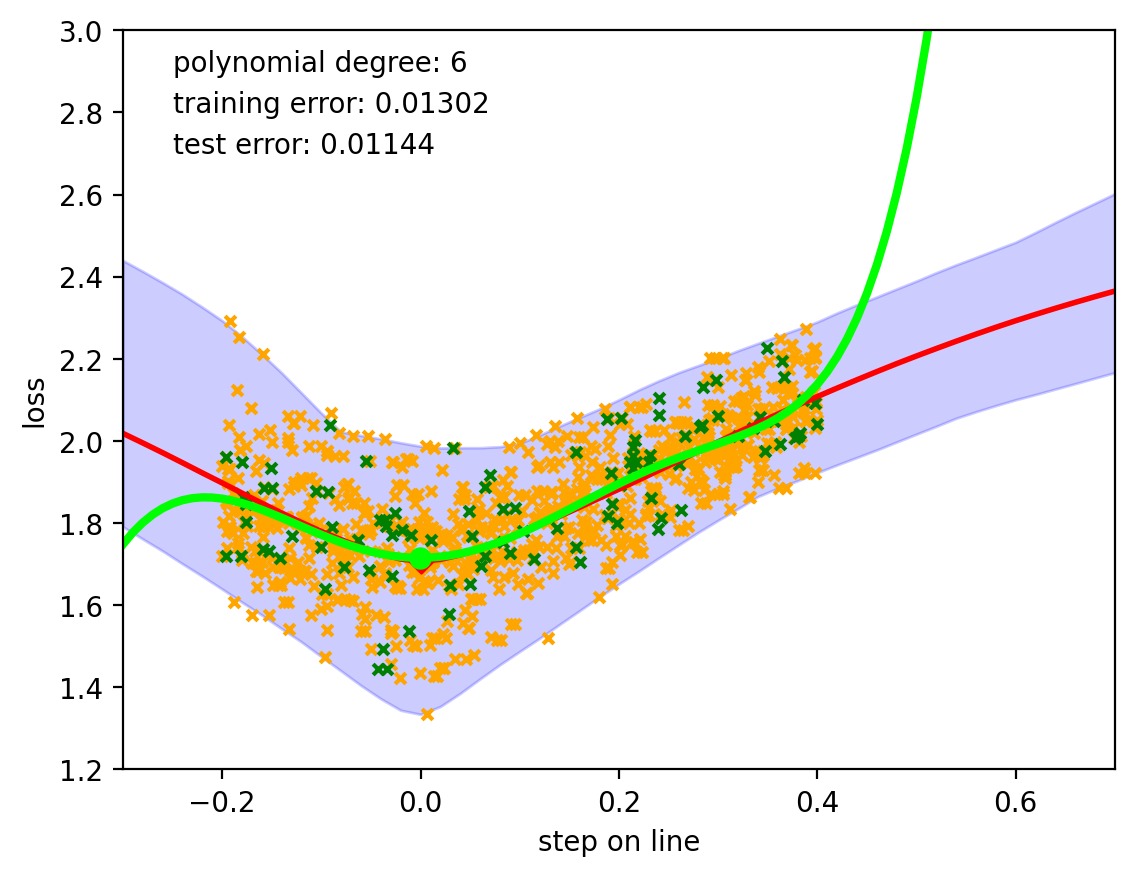}
	\includegraphics[width=\disfigwidth\linewidth]{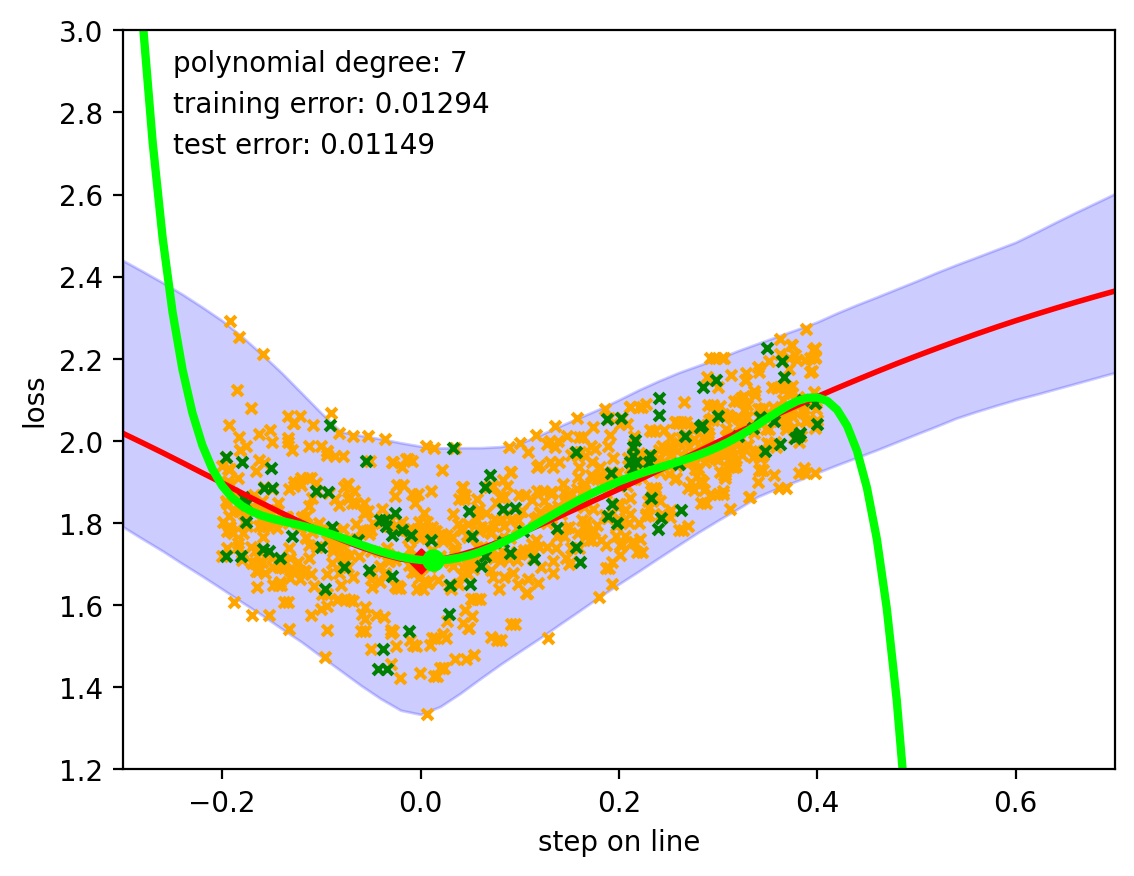}\\
	\vspace{-0.1cm}
	%\vspace{-0.2cm}
	%\includegraphics[width=\disfigwidth\linewidth]{plot_data/distribution_plots/train_line306}
	%\includegraphics[width=\disfigwidth\linewidth]{plot_data/distribution_plots/train_line307}
	%\includegraphics[width=\disfigwidth\linewidth]{plot_data/distribution_plots/train_line308}
\label{fig:polynomial_fits}
\caption{An exemplar ELF routine testing for the best fitting polynomial. The empirical loss in given in \textcolor{red}{red}, the distribution of batch losses in \textcolor{blue}{blue}. The sampled losses are given in \textcolor{YellowOrange}{orange} and \textcolor{ForestGreen}{green}. The green losses are the test set of the current cross validation step. It can be seen, that the fifth-order polynomial (\textcolor{green}{green}) reaches the lowest test error.} 
\vspace{-0.5cm}
\end{figure}

Our straightforward concept is to sample $n$ losses $\mathcal{L}_{batch}(\mathbf{\theta}_0+s_i\cdot-\hat{\nabla}_\mathbf{\theta}\mathcal{L}_{batch}(\mathbf{\theta},\mathbb{B}_0),\mathbb{B}_i)$, with $i$ ranging from $1$ to $n$ and $\mathbb{B}_i$ uniformly chosen from $\mathbb{T}$ and $s_i$ uniformly chosen from a reasonable interval, on which we will focus later. Now, we follow a classical function fitting or machine learning routine.
An ordinary least square regression  (OLSR) for polynomials is performed. Note, that our data is not homoscedastic, as required for OLSR \footnote{We indirectly use weighted OLSR by sampling more points in  relevant intervals around the minimum, which softens the effect of heteroscedasticity.}. This implies, that our resulting estimator is still unbiased, but we cannot perform an analysis of variances (see \cite{goldberger1964econometric}). However, the latter is not needed in our case.  Those regressions are performed with increasing degree until the test error of the fitted polynomial is increasing. The test error is determined by a 5-fold cross-validation. The second last polynomial degree is chosen and the polynomial is again fitted on all loss values to get a more accurate fit. Consequently the closest minimum to the initial location is determined and additional losses are measured in a reasonable interval around it. This process is repeated four times. Finally, the step to the closest minimum of the fitted polynomial is chosen as update step if, existing and its value is positive. Otherwise, a new line search is started. This routine is described in more detail in Algorithm \ref{algorithm:ELF_line_search}. An empirical example of the search of the best fitting polynomial is given in Figure \ref{fig:polynomial_fits}. The empirically found heuristic to determine reasonable measure intervals is given in Algorithm \ref{algorithm:chose_interval}. This routine empirically ensures, that the point cloud of losses is wider than high, so that a correctly oriented polynomial is fitted.  
To determine when to measure a new step size with a line search, we utilize that one can estimate the expected improvement by $l_{ELF}(0)-l_{ELF}(s_{min})$. If the real improvement of the training loss times a factor is smaller than the expected improvement, both determined over a step window, a new step size is determined. 
The full ELF algorithm is given in Algorithm \ref{algorithm:elf} in Appendix \ref{app:elf}. 
We note that all of the mentioned subroutines are easy to implement with the Numpy Python library, which reduces the implementation effort significantly. The presented pseudo codes include the most important aspects for understanding our approach. For a more detailed description we refer to our implementation found in the supplementary material.\\
Based on our empirical experience with this approach we introduce the following additions:
\textbf{1.} We measure 3 consecutive lines and take the average resulting step size to continue training with SGD.
\textbf{2.} We have experienced that ELF generalizes better if not performing a step to the minimum, but to perform a step that decreased the loss by a decrease factor $\delta$ of the overall improvement. In detail, we estimate $x_{target}> x_{min}$, which satisfies  $f(x_{target})=\delta (f(x_0)-x_{min})-f(x_{min})$ with $\delta \in [0,1)$
\textbf{3.} We use a momentum term $\beta$ on the gradient, which can lead to an improvement in generalization. \textbf{4.} To prevent over-fitting, the batch losses required for fitting polynomials are sampled from the validation set. \textbf{5.} At the beginning of the training a short grid search is done to find the maximal step size that still supports training. This reduces the chances of getting stuck in a local minima at the beginning of optimization.
% The reason might be that we are now not directly moving to the nearest minimum?.

\begin{algorithm}[t!]
\caption{Pseudo code of ELF's line search routine (see Algorithm \ref{algorithm:elf})}
\begin{algorithmic}[1]
\label{algorithm:ELF_line_search}
\renewcommand{\algorithmicrequire}{\textbf{Input:}}
\newcommand{\algorithmicbreak}{\textbf{break}}
\newcommand{\BREAK}{\STATE \algorithmicbreak}

\REQUIRE $\mathbf{d}$: direction (default: current unit gradient)
\REQUIRE $\mathbf{\theta}_0$: initial parameter space position
\REQUIRE $\mathcal{L}_{batch}(\mathbf{\theta}_t)$: batch loss
function which randomly chooses a batch
\REQUIRE $k$: sample interval adaptations (default: 5)
\REQUIRE $n$: losses to sample per adaptation (default: 100)

%	\ENSURE 
%	\\ \textit{Initialisation} :
%<put decay term or nor?
\STATE interval\_width $\leftarrow$ 1.0
\STATE sample\_positions $\leftarrow$ []
\STATE lineLosses $\leftarrow$ []
\FOR{$r$ from 0 to k}
	\IF{r != 0}
	\STATE interval\_width $\leftarrow$ chose\_sample\_interval(minimum\_location, sample\_positions, line\_losses, coefficents) 
	\ENDIF
	\STATE new\_sample\_positions $\leftarrow$ get\_uniformly\_distributed\_values(n, interval\_width)
	\FOR{$m$ in new\_sample\_positions}
		\STATE line\_losses.append($\mathcal{L}_{batch}(\mathbf{\theta}_0+m \mathbf{d})$)
	\ENDFOR
	\STATE sample\_positions.extend(new\_sample\_positions)
	\STATE last\_test\_error $\leftarrow$ $\infty$
	\FOR{degree from 0 to max\_polynomial\_degree}
		\STATE test\_error $\leftarrow$ 5-fold\_cross\_validation(degree, sample\_positions, line\_losses)
		\IF{last\_test\_error $<$ test\_error}
			\STATE best\_degree $\leftarrow$ degree$-1$
			\STATE last\_test\_error $\leftarrow$ test\_error
			\BREAK
		\ENDIF
		\IF{degree $==$ max\_polynomial\_degree}
			\STATE best\_degree $\leftarrow$ max\_polynomial\_degree
			\BREAK
		\ENDIF
	\ENDFOR
	\STATE coefficients $\leftarrow$ fit\_polynomial(best\_degree, sample\_positions, line\_losses)
	\STATE minimum\_position, improvement $\leftarrow$ get\_minimum\_position\_nearest\_to\_0(coefficients)
	\ENDFOR
\RETURN minimum\_position, improvement, $k\cdot n$
\end{algorithmic} 
\end{algorithm}

\begin{algorithm}[b!]
\caption{Pseudo code of the chose\_sample\_interval routine of Algorithm \ref{algorithm:ELF_line_search}}
\begin{algorithmic}[1]
\label{algorithm:chose_interval}
\renewcommand{\algorithmicrequire}{\textbf{Input:}}
\newcommand{\algorithmicbreak}{\textbf{break}}
\newcommand{\BREAK}{\STATE \algorithmicbreak}

\REQUIRE minimum\_position 
\REQUIRE sample\_positions 
\REQUIRE line\_losses : list of losses corresponding to the sample\_positions
\REQUIRE coefficents : coefficients of the polynomial of the current fit
\REQUIRE min\_window\_size (default: 50)

\STATE window $\leftarrow$ $\{m : m \in \text{sample\_positions}  $ and $ 0\leq m \leq 2 \cdot \text{minimum\_location}\}$
\IF{$|window| <$ min\_window\_size}
	\STATE window $\leftarrow$ get\_50\_nearest\_sample\_pos\_to\_minimum\_pos-\\
	\hspace{1.6cm}(sample\_positions, minimum\_position)
\ENDIF
\STATE target\_loss $\leftarrow$ get\_third\_quartile(window, line\_losses[window])
\STATE interval\_width $\leftarrow$ get\_nearest\_position\_where\_the\_absolut\_of\_polynomial-\\ \hspace{1.6cm}\_takes\_value(coefficents,target\_loss)
\RETURN interval\_width
\end{algorithmic} 
\end{algorithm}

\section{Empirical Analasys}
To make our analysis comparable on steps and epochs, we define one step as loading of a new input batch. Thus, the steps/batches needed of ELF to estimate a new line step are included.
\vspace{-0.25cm}
\subsection{Comparison to optimization approaches operating on $\mathcal{L}_{emp}$}
\label{subsec:emplosscomparison}
				\begin{figure}[t!]
				\vspace{-0.7cm}
					\label{fig:elf_comp}
						\newcommand\picscale{0.3}
						\newcommand\stepmin{1E-3}
						\newcommand\stepemax{25.0}
						\newcommand\steps{140E3}
						\newcommand\lossmin{0.0035}
						\newcommand\lossmax{1}
						\newcommand\evalmin{0.5}
						\newcommand\evalmax{1.0}
						\newcommand\epochs{300}
						
						%%%
						% line search
						%%%
						\begin{tikzpicture}[scale=\picscale] % ccoordinate scale , does not affect text
						\begin{axis}[
						ymode=log,
						width=\linewidth, % Scale the plot to \linewidth
						grid=major, % Display a grid
						grid style={dashed,gray!30}, % Set the style, 
						xlabel=training step, % Set the labels
						ylabel=Step Size,
						xmin=0,xmax=\steps,
						ymin=\stepmin,ymax=\stepemax,
						p1/.style={draw=ForestGreen,line width=2pt},
						p2/.style={draw=blue,line width=2pt},%mark=*,mark options={fill=white}}
						p3/.style={draw=red,line width=3.5pt},%mark=*,mark options={fill=white}}
						title=Step Size,
						title style={font=\Huge},
						title style={font=\huge},
						label style={font=\LARGE},
						x tick label style={rotate=0,anchor=near xticklabel,font=\Large}, 
						y tick label style={font=\Large}, 
						]
						\addplot [p1] table[x=Step,y=Value,col sep=comma] {plot_data/lemp_comparison/golsi_step_size.csv}; 
						\addplot [p2] table[x=Step,y=Value,col sep=comma] {plot_data/lemp_comparison/pls_step_size.csv};  
						\addplot [p3] table[x=Step,y=Value,col sep=comma] {plot_data/lemp_comparison/elf_step_size.csv};
						\end{axis}
						\end{tikzpicture}
						\begin{tikzpicture}[scale=\picscale] % ccoordinate scale , does not affect text
						\begin{axis}[
						width=\linewidth, % Scale the plot to \linewidth
						grid=major, % Display a grid
						grid style={dashed,gray!30}, % Set the style, transperancz
						xlabel=epoch, % Set the labels
						ylabel=Validation Accuracy,
						xmin=0,xmax=\epochs,
						ymin=\evalmin,ymax=\evalmax,
						legend style={at={(0.75,0.25)},anchor=north,font=\huge}, % Put the legend left of the plot
						%legend style=={cells={align=left}}, % Put the legend left of the plot
						p1/.style={draw=ForestGreen,line width=2.5pt},
						p2/.style={draw=blue,line width=2.5pt},%mark=*,mark options={fill=white}}
						p3/.style={draw=red,line width=2.5pt},%mark=*,mark options={fill=white}}
						p4/.style={draw=black,line width=2.5pt},%mark=*,mark options={fill=white}}
						title=Validation Accuracy,
						title style={font=\huge},
						label style={font=\LARGE},
						x tick label style={rotate=0,anchor=near xticklabel,font=\Large}, 
						y tick label style={font=\Large}, 
						]
						\addplot [p1] table[x=Step,y=Value,col sep=comma] {plot_data/lemp_comparison/golsi_val_acc.csv}; 
						\addplot [p2] table[x=Step,y=Value,col sep=comma] {plot_data/lemp_comparison/pls_val_acc.csv}; 
						\addplot [p3] table[x=Step,y=Value,col sep=comma] {plot_data/lemp_comparison/elf_val_acc.csv}; 
						\legend{GOLS1,PLS,ELF}
						\end{axis}
						\end{tikzpicture}
						\begin{tikzpicture}[scale=\picscale] % ccoordinate scale , does not affect text
						\begin{axis}[
						ymode=log,
						width=\linewidth, % Scale the plot to \linewidth
						grid=major, % Display a grid
						grid style={dashed,gray!30}, % Set the style, transperancz
						xlabel=epoch, % Set the labels
						ylabel=Training Loss,
						xmin=0,xmax=\epochs,
						ymin=\lossmin,ymax=\lossmax,
						legend style={at={(-0.5,0.5)},anchor=north}, % Put the legend left of the plot
						p1/.style={draw=ForestGreen,line width=2.5pt},
						p2/.style={draw=blue,line width=2.5pt},%mark=*,mark options={fill=white}}
						p3/.style={draw=red,line width=2.5pt},%mark=*,mark options={fill=white}}
						p4/.style={draw=black,line width=2.5pt},%mark=*,mark options={fill=white}}
						title=Training Loss,
						title style={font=\huge},
						label style={font=\LARGE},
						x tick label style={rotate=0,anchor=near xticklabel,font=\Large}, 
						y tick label style={font=\Large}, 
						]
						\addplot [p1] table[x=Step,y=Value,col sep=comma] {plot_data/lemp_comparison/golsi_train_loss.csv}; 
						\addplot [p2] table[x=Step,y=Value,col sep=comma] {plot_data/lemp_comparison/pls_train_loss.csv};
						\addplot [p3] table[x=Step,y=Value,col sep=comma] {plot_data/lemp_comparison/elf_train_loss.csv};
						\end{axis} 
						\end{tikzpicture}
						\caption{Comparison of ELF against PLS and GOLS1 on a ResNet-32 trained on CIFAR-10. The corresponding test accuracies are: EFL: 0.900, PLS: 0.875, GOLS1: 0.872. ELF performs better in this scenario and, intriguingly, PLS and ELF estimate similar step size schedules. }
				\end{figure}
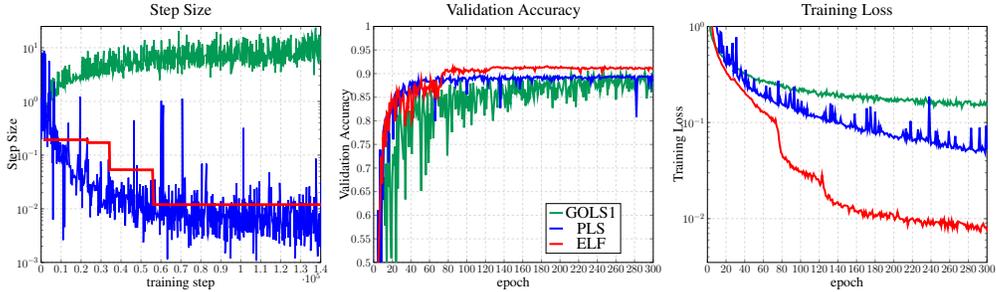

We compare against PLS \citep{probabilisticLineSearch} and GOLS1 \citep{gradientOnlyLineSearch}. Both approximate line searches on the empirical loss. Since PLS has to be adapted manually for each Model that introduces new layers, we restrict our comparison to a ResNet-32 (\cite{resnet}) trained on CIFAR-10 (\cite{CIFAR-10}). In addition, we use an empirically optimized and the only available Tensorflow (\cite{Tensorflow}) implementation of PLS (\cite{probabilisticLineSearchImpl}). For each optimizer we tested five appropriate hyperparameter combinations, which are likely to result in good results (see Appendix \ref{app:hyperparam_emp_loss_comp}).
The best performing runs are given in Figure \ref{fig:elf_comp}. We can see that ELF slightly surpasses GOLS1 and PLS on validation accuracy and training loss in this scenario. Intriguingly, PLS and ELF estimate similar step size schedules, whereas, that of GOLS1 differs significantly.

\subsection{Robustness comparison of ELF, ADAM and SGD}
\label{subsec:robustness}
\begin{figure}[t!]
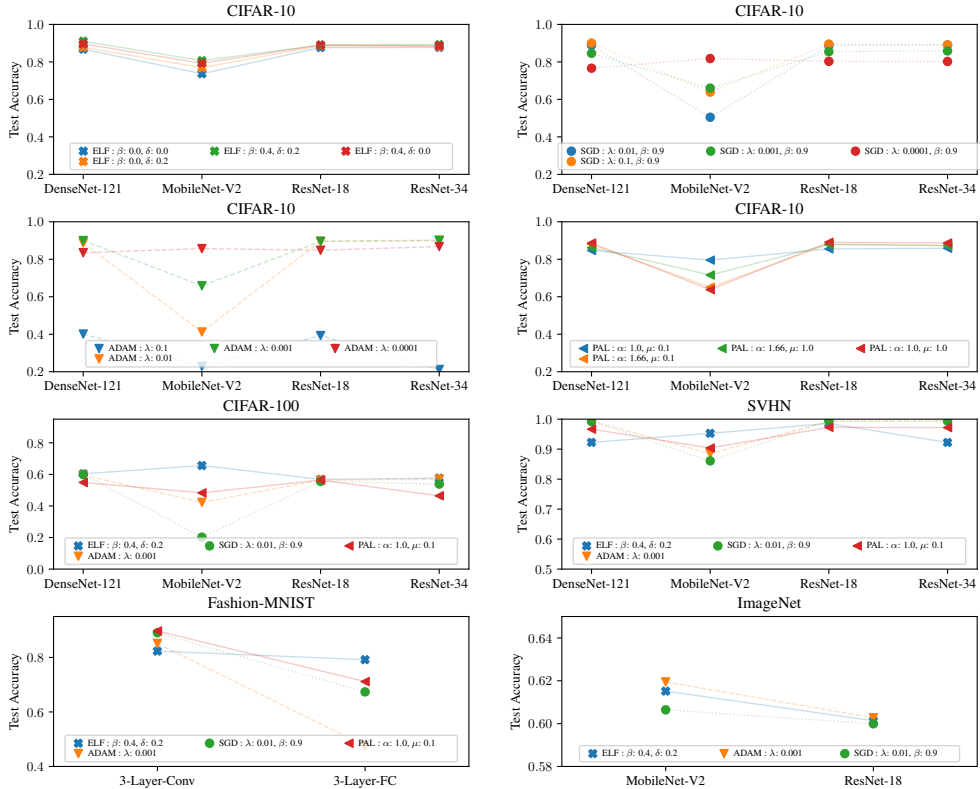

\centering
%\scalebox{0.51}{
%\input{../../../PycharmProjects/LineSearchExperiments_Pytorch/Evaluation/optimizer_comparison/plots/CIFAR-10_ELF.pgf}}\hspace{-0.5cm}
%\scalebox{0.51}{
%\input{../../../PycharmProjects/LineSearchExperiments_Pytorch/Evaluation/optimizer_comparison/plots/CIFAR-10_SGD.pgf}}\\
%\scalebox{0.51}{
%\input{../../../PycharmProjects/LineSearchExperiments_Pytorch/Evaluation/optimizer_comparison/plots/CIFAR-10_ADAM.pgf}}\hspace{-0.5cm}
%\scalebox{0.51}{
%\input{../../../PycharmProjects/LineSearchExperiments_Pytorch/Evaluation/optimizer_comparison/plots/CIFAR-10_PAL.pgf}}\\
%\scalebox{0.51}{
%\input{../../../PycharmProjects/LineSearchExperiments_Pytorch/Evaluation/optimizer_comparison/plots/CIFAR-100.pgf}}\hspace{-0.5cm}
%\scalebox{0.51}{
%\input{../../../PycharmProjects/LineSearchExperiments_Pytorch/Evaluation/optimizer_comparison/plots/SVHN.pgf}}
%\scalebox{0.51}{
%\input{../../../PycharmProjects/LineSearchExperiments_Pytorch/Evaluation/optimizer_comparison/plots/Fashion-MNIST.pgf}}\hspace{-0.5cm}
%\scalebox{0.51}{
%\input{../../../PycharmProjects/LineSearchExperiments_Pytorch/Evaluation/optimizer_comparison/plots/ImageNet.pgf}}
\scalebox{0.51}{
\input{plot_data/optimizer_comparison/plots/CIFAR-10_ELF.pgf}}\hspace{-0.5cm}
\scalebox{0.51}{
\input{plot_data/optimizer_comparison/plots/CIFAR-10_SGD.pgf}}\\
\scalebox{0.51}{
\input{plot_data/optimizer_comparison/plots/CIFAR-10_ADAM.pgf}}\hspace{-0.5cm}
\scalebox{0.51}{
\input{plot_data/optimizer_comparison/plots/CIFAR-10_PAL.pgf}}\\
\scalebox{0.51}{
\input{plot_data/optimizer_comparison/plots/CIFAR-100.pgf}}\hspace{-0.5cm}
\scalebox{0.51}{
\input{plot_data/optimizer_comparison/plots/SVHN.pgf}}
\scalebox{0.51}{
\input{plot_data/optimizer_comparison/plots/Fashion-MNIST.pgf}}\hspace{-0.5cm}
\scalebox{0.51}{
\input{plot_data/optimizer_comparison/plots/ImageNet.pgf}}

\caption{ Robustness comparison of ELF, ADAM , SGD and PAL. The most robust optimizer-hyperparameters across several models were determined on CIFAR-10, which then tested on further datasets and models. On those, the found hyperparameters of ELF, ADAM, SGD and PAL behave robust. ($\beta$=momentum, $\lambda$=learning rate, $\delta$=decrease factor,$\alpha$=update step adaptation, $\mu$=measuring step size). Plots of the training loss and validation accuracy are given in Appendix \ref{app:train_eval_plots}. (Results for PAL on ImageNet will be included in the final version.)}
\label{fig:robustness_comparison}
\end{figure}

\begin{figure}[b!]
\centering
\includegraphics[width=0.24\linewidth]{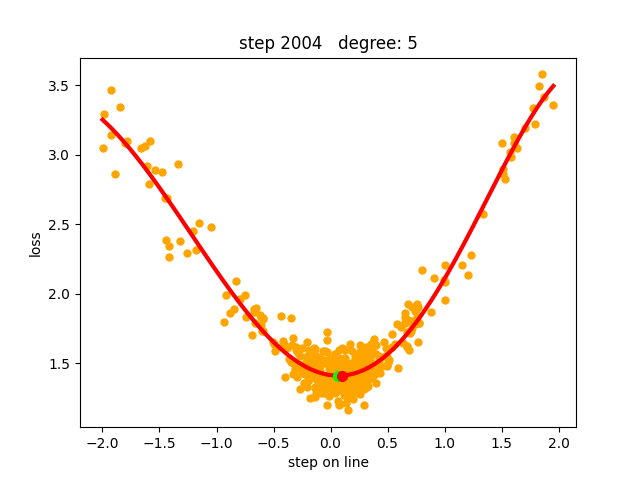}
\includegraphics[width=0.24\linewidth]{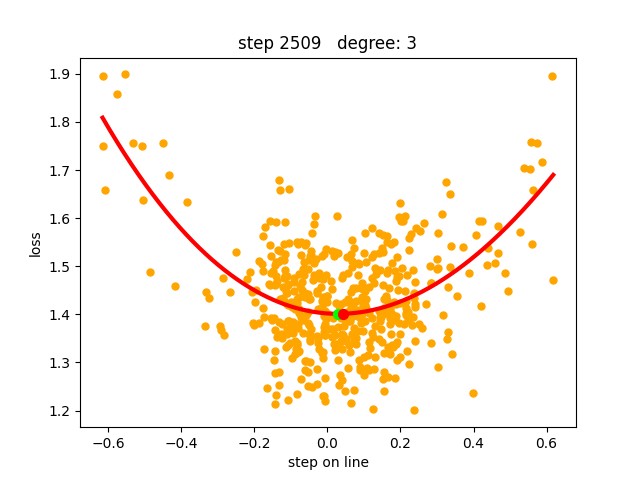}
\includegraphics[width=0.24\linewidth]{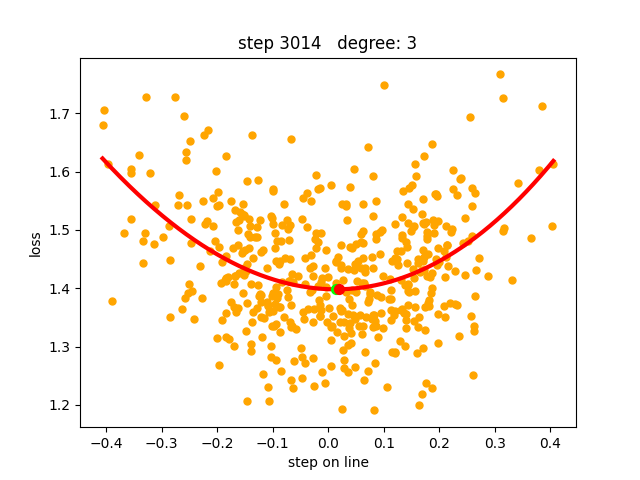}
\includegraphics[width=0.24\linewidth]{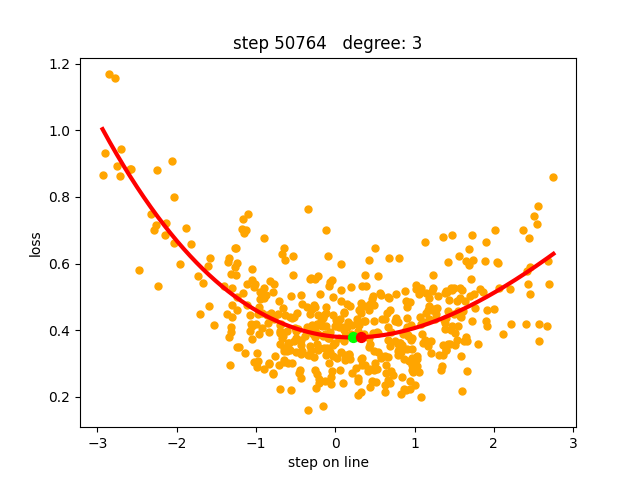}
\caption{Representative polynomial line approximations (\textcolor{red}{red}) obtained by training ResNet-18 on CIFAR-10. The samples losses are depicted in \textcolor{YellowOrange}{orange}. The minimum of the approximation is represented by the \textcolor{green}{green dot}, whereas the update step adjusted by a decrease factor of 0.2 is depicted as the  \textcolor{red}{red dot}.}
\label{fig:sampledlines}
\end{figure}

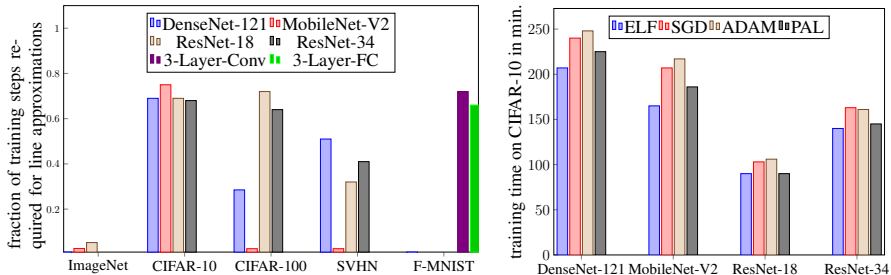
\begin{figure}[t!]
\centering
\begin{tikzpicture}[scale=0.4]
\begin{axis}[
	width= 1.1\linewidth,
	height= 0.7\linewidth,
	ylabel= fraction of training steps required for line approximations,
	legend style={at={(0.5,0.97)},
	anchor=north,legend columns=-1},
	ybar ,
	symbolic x coords={ImageNet,CIFAR-10,CIFAR-100,SVHN,F-MNIST},
	ylabel style={text width=8cm,font=\LARGE},
	xlabel style={font=\LARGE},
	x tick label style={font=\Large},
	%y tick label style={font=\Large},
	legend style={font=\LARGE},
	xtick=data,
	xtick pos=left,
	ytick pos=left,
	ymin=0.01,ymax=1.1,
	legend columns=2,
]
\addplot 
	coordinates {(CIFAR-10,0.69) (CIFAR-100,0.285) (SVHN,0.51) (F-MNIST,0) (ImageNet,0)};
\addplot 
	coordinates {(CIFAR-10,0.75) (CIFAR-100,0.025) (SVHN,0.025) (ImageNet,0.026)};
%\addplot 
%	coordinates {(CIFAR-10,0.75) (CIFAR-100,0.01) (SVHN,0.01) (ImageNet,0.016)}; correct values but not visisble
\addplot 
	coordinates {(CIFAR-10,0.69) (CIFAR-100,0.72) (SVHN,0.32) (ImageNet,0.052)};
\addplot 
	coordinates {(CIFAR-10,0.68) (CIFAR-100,0.64) (SVHN,0.41)};
\addplot 
	coordinates {(F-MNIST,0.72)};
\addplot 
	coordinates {(F-MNIST,0.66)};

\legend{DenseNet-121,MobileNet-V2, ResNet-18, ResNet-34,3-Layer-Conv,3-Layer-FC}
\end{axis}
\end{tikzpicture}\quad
\begin{tikzpicture}[scale=0.4]
\begin{axis}[
	width= 0.9\linewidth,
	height= 0.7\linewidth,
	ylabel= training time on CIFAR-10 in min.,
	legend style={at={(0.5,0.97)},
	anchor=north,legend columns=-1},
	ybar ,
	symbolic x coords={DenseNet-121,MobileNet-V2,ResNet-18,ResNet-34},
	ylabel style={font=\LARGE},
	xlabel style={font=\LARGE},
	x tick label style={font=\Large},
	y tick label style={font=\Large},
	legend style={font=\LARGE},
	xtick=data,
	xtick pos=left,
	ytick pos=left,
	ymin=0.0,%ymax=0.87,
]
\addplot 
	coordinates {(DenseNet-121,207) (MobileNet-V2,165) (ResNet-18,90) (ResNet-34,140)};
\addplot 
	coordinates {(DenseNet-121,240) (MobileNet-V2,207) (ResNet-18,103) (ResNet-34,163)};
\addplot 
	coordinates {(DenseNet-121,248) (MobileNet-V2,217) (ResNet-18,106) (ResNet-34,161)};
\addplot 
	coordinates {(DenseNet-121,225) (MobileNet-V2,186) (ResNet-18,90) (ResNet-34,145)};

\legend{ELF,SGD,ADAM,PAL}
\end{axis}
\end{tikzpicture}
\caption{Left: Fraction of training steps used for step size estimation and the total amount of training steps. The amount of steps needed depends strongly on the model and dataset. Right: Training time comparison on CIFAR-10. One can observe, that ELF performs slightly faster. }
\label{fig:speed_comp}
\end{figure}

Since ELF, as long as loss lines are well approximate-able by polynomials, should adapt to different loss landscapes, it is expected to perform robustly across models and datasets. Therefore, we will concentrate on robustness in this evaluation. For CIFAR-10, CIFAR-100, SVHN we will train on DenseNet-121, MobileNetV2, ResNet-18 and ResNet-34. For Fashion-MNIST we consider a 3-layer fully connected network and a 3-layer convolutional network. For ImageNet we consider ResNet-18 and MobileNetV2. We compare against the most widely used optimizer SGD, ADAM (widely considered to be robust \citep{schmidt2020descending,adam}) and PAL (\cite{pal}), a line search approach on $\mathcal{L}_{batch}$. At first, we perform an optimizer-hyperparameter grid search over the models considered for CIFAR-10. Then, we choose the hyperparameter combination that achieves on average the best test accuracy for each optimizer. Consequently robustness is evaluated by reusing those hyperparameters on each additional dataset and models. 
For ADAM and SGD we use a standard learning rate schedule, which divides the initial learning rate by 10 after half and after three quarters of the training steps. For ADAM and SGD we considered $10^{-1},10^{-2},10^{-3},10^{-4}$ as learning rates. ADAM's moving average factors are set to $0.9$ and $0.999$. For SGD, we used a momentum of $0.9$.
For ELF we used the default value for each hyperparameter, except for the momentum factor, which was chosen as $0$ or $0.4$ and the decrease factor $\delta$ was chosen as $0$ or $0.2$. For PAL a measuring step size of $0.1$ or $1$ and a step size adaptation of $1.66$ and $1.0$ was considered. Further details are given in Appendix \ref{app:hyperparam_emp_loss_comp}.\\
As shown in Figure \ref{fig:robustness_comparison} our experiments revealed that the most robust hyperparameter combination is a momentum factor of $0.4$ and a $\delta$ of 0.2 for ELF. For SGD, the most robust learning rate is $0.01$  and for ADAM $0.001$. For PAL a measuring step size of $0.1$  and a step size adaptation $1.0$ perform most robustly. 

For all optimizers the most robust hyperparameters found on CIFAR-10 tend to perform also robustly on other datasets and models. This is unexpected, since it is generally assumed that a new learning rate has to be searched for each new problem.
In addition, we note that ELF tends to perform better on MobileNet-V2 and the 3-Layer-FC network, however, is not that robust on SVHN. 
The most important insight, which has been obtained from our experiments, is that for all tested models and dataset ELF was able to fit lines by polynomials. Thus, we are able to directly measure step sizes  on the empirical loss. Furthermore, we see that those step sizes are useful to guide optimization on subsequent steps. To strengthen this statement, we plotted the sampled losses and the fitted polynomials for each approximated line. Representative examples are given in Figure \ref{fig:sampledlines} and in Appendix \ref{app:furtherLineApp}.\\
Figure \ref{fig:speed_comp} (left) shows that ELF uses depending on the models between $1\%$ to $75\%$ of its training steps to approximate lines. This shows, that depending on the problem more or less step sizes have to be determined. In addition, this indicates, that update steps become more efficient for models, for which many new step sizes had to be determined.
Figure \ref{fig:speed_comp} (left) shows that ELF is faster than the other optimizers. This is a consequence of sole forward passes required for measuring the losses on lines and since the operations required to fit polynomials are cheap.

\section{Discussion \& Outlook }
This work demonstrates that line searches on the true empirical loss can be done efficiently with the help of polynomial approximations. These approximations are valid for all investigated models and datasets. Although, from a practical point of view, measuring a new step size is still expensive, it seems to be sufficient for a successful training process to measure only a few exact step sizes during training. Our optimization approach ELF  performs robustly over datasets and models without any hyperparameter tuning needed and competes against PAL, ADAM and SGD. The later 2 were run with a good learning rate schedule.
In our experiments ELF showed better performance than Probabilistic Line Search and Gradient-only Line Search. Both also estimate their update steps from the expected empirical loss.

An open question is to what extent our approach leads to improvements in theory. It is known that exact steepest descent line searches on deterministic problems (\cite{nonlinear_programming}, p. 239)) exhibit linear convergence on any function, which is twice continuously differentiable and has a relative minimum at which the hessian is positive definite. The question to be considered is how the convergence behavior of an exact line search behaves in noisy steepest directions. This, to the best of our knowledge, has not yet been answered.

%\begin{tikzpicture}
%\begin{axis}[
%	width= 0.9\linewidth,
%	ylabel= \# approximated lines,
%	enlargelimits=0.2,
%	legend style={at={(0.5,-0.15)},
%	anchor=north,legend columns=-1},
%	ybar ,
%	symbolic x coords={CIFAR-10,CIFAR-100,SVHN,F-MNIST},
%	xtick=data,
%]
%\addplot 
%	coordinates {(CIFAR-10,46) (CIFAR-100,19) (SVHN,34) (F-MNIST,0)};
%\addplot 
%	coordinates {(CIFAR-10,75) (CIFAR-100,1) (SVHN,1)};
%\addplot 
%	coordinates {(CIFAR-10,69) (CIFAR-100,72) (SVHN,32)};
%\addplot 
%	coordinates {(CIFAR-10,68) (CIFAR-100,64) (SVHN,41)};
%\addplot 
%	coordinates {(F-MNIST,48)};
%\addplot 
%	coordinates {(F-MNIST,44)};
%
%\legend{DenseNet-121,MobileNet-V2, ResNet-18, ResNet-34,3-Layer-Conv,3-Layer-FC}
%\end{axis}
%\end{tikzpicture}

%\bibliography{iclr2021_conference}
%\bibliographystyle{iclr2021_conference}
\vfill
\pagebreak
{%\small
	%\nocite{*}
\bibliographystyle{iclr2021_conference}
\bibliography{iclr2021_conference}

\begin{thebibliography}{30}
\providecommand{\natexlab}[1]{#1}
\providecommand{\url}[1]{\texttt{#1}}
\expandafter\ifx\csname urlstyle\endcsname\relax
  \providecommand{\doi}[1]{doi: #1}\else
  \providecommand{\doi}{doi: \begingroup \urlstyle{rm}\Url}\fi

\bibitem[Abadi et~al.(2015)Abadi, Agarwal, Barham, Brevdo, Chen, Citro,
  Corrado, Davis, Dean, Devin, Ghemawat, Goodfellow, Harp, Irving, Isard, Jia,
  Jozefowicz, Kaiser, Kudlur, Levenberg, Man\'{e}, Monga, Moore, Murray, Olah,
  Schuster, Shlens, Steiner, Sutskever, Talwar, Tucker, Vanhoucke, Vasudevan,
  Vi\'{e}gas, Vinyals, Warden, Wattenberg, Wicke, Yu, and Zheng]{Tensorflow}
Mart\'{\i}n Abadi, Ashish Agarwal, Paul Barham, Eugene Brevdo, Zhifeng Chen,
  Craig Citro, Greg~S. Corrado, Andy Davis, Jeffrey Dean, Matthieu Devin,
  Sanjay Ghemawat, Ian Goodfellow, Andrew Harp, Geoffrey Irving, Michael Isard,
  Yangqing Jia, Rafal Jozefowicz, Lukasz Kaiser, Manjunath Kudlur, Josh
  Levenberg, Dandelion Man\'{e}, Rajat Monga, Sherry Moore, Derek Murray, Chris
  Olah, Mike Schuster, Jonathon Shlens, Benoit Steiner, Ilya Sutskever, Kunal
  Talwar, Paul Tucker, Vincent Vanhoucke, Vijay Vasudevan, Fernanda Vi\'{e}gas,
  Oriol Vinyals, Pete Warden, Martin Wattenberg, Martin Wicke, Yuan Yu, and
  Xiaoqiang Zheng.
\newblock {TensorFlow}: Large-scale machine learning on heterogeneous systems,
  2015.
\newblock URL \url{https://www.tensorflow.org/}.
\newblock Software available from tensorflow.org.

\bibitem[Balles(2017)]{probabilisticLineSearchImpl}
Lukas Balles.
\newblock Probabilistic line search tensorflow implementation, 2017.
\newblock URL
  \url{https://github.com/ProbabilisticNumerics/probabilistic_line_search/commit/a83dfb0}.

\bibitem[Baydin et~al.(2017)Baydin, Cornish, Rubio, Schmidt, and
  Wood]{hypergradientdescent}
Atilim~Gunes Baydin, Robert Cornish, David~Martinez Rubio, Mark Schmidt, and
  Frank Wood.
\newblock Online learning rate adaptation with hypergradient descent.
\newblock \emph{arXiv preprint arXiv:1703.04782}, 2017.

\bibitem[Berahas et~al.(2019)Berahas, Jahani, and Tak{\'a}c]{S-LBFGS}
Albert~S. Berahas, Majid Jahani, and Martin Tak{\'a}c.
\newblock Quasi-newton methods for deep learning: Forget the past, just sample.
\newblock \emph{CoRR}, abs/1901.09997, 2019.

\bibitem[Berrada et~al.(2019)Berrada, Zisserman, and Kumar]{L4_alternative}
Leonard Berrada, Andrew Zisserman, and M~Pawan Kumar.
\newblock Training neural networks for and by interpolation.
\newblock \emph{arXiv preprint arXiv:1906.05661}, 2019.

\bibitem[Botev et~al.(2017)Botev, Ritter, and Barber]{gausnewton}
Aleksandar Botev, Hippolyt Ritter, and David Barber.
\newblock Practical gauss-newton optimisation for deep learning.
\newblock \emph{arXiv preprint arXiv:1706.03662}, 2017.

\bibitem[Chae \& Wilke(2019)Chae and Wilke]{empericalLineSearchApproximations}
Younghwan Chae and Daniel~N Wilke.
\newblock Empirical study towards understanding line search approximations for
  training neural networks.
\newblock \emph{arXiv preprint arXiv:1909.06893}, 2019.

\bibitem[Goldberger et~al.(1964)]{goldberger1964econometric}
Arthur~Stanley Goldberger et~al.
\newblock Econometric theory.
\newblock \emph{Econometric theory.}, 1964.

\bibitem[Goodfellow et~al.(2016)Goodfellow, Bengio, Courville, and
  Bengio]{goodfellow2016deep}
Ian Goodfellow, Yoshua Bengio, Aaron Courville, and Yoshua Bengio.
\newblock \emph{Deep learning}, volume~1.
\newblock MIT press Cambridge, 2016.

\bibitem[He et~al.(2016)He, Zhang, Ren, and Sun]{resnet}
Kaiming He, Xiangyu Zhang, Shaoqing Ren, and Jian Sun.
\newblock Deep residual learning for image recognition.
\newblock In \emph{Proceedings of the IEEE conference on computer vision and
  pattern recognition}, pp.\  770--778, 2016.

\bibitem[Jorge~Nocedal(2006)]{numerical_optimization}
Stephen~Wright Jorge~Nocedal.
\newblock \emph{Numerical Optimization}.
\newblock Springer series in operations research. Springer, 2nd ed edition,
  2006.
\newblock ISBN 9780387303031,0387303030.

\bibitem[Kafka \& Wilke(2019)Kafka and Wilke]{gradientOnlyLineSearch}
Dominic Kafka and Daniel Wilke.
\newblock Gradient-only line searches: An alternative to probabilistic line
  searches.
\newblock \emph{arXiv preprint arXiv:1903.09383}, 2019.

\bibitem[Kingma \& Ba(2014)Kingma and Ba]{adam}
Diederik~P. Kingma and Jimmy Ba.
\newblock Adam: {A} method for stochastic optimization.
\newblock \emph{CoRR}, abs/1412.6980, 2014.
\newblock URL \url{http://arxiv.org/abs/1412.6980}.

\bibitem[Krizhevsky \& Hinton(2009)Krizhevsky and Hinton]{CIFAR-10}
Alex Krizhevsky and Geoffrey Hinton.
\newblock Learning multiple layers of features from tiny images.
\newblock Technical report, Citeseer, 2009.

\bibitem[Liu et~al.(2019)Liu, Jiang, He, Chen, Liu, Gao, and Han]{radam}
Liyuan Liu, Haoming Jiang, Pengcheng He, Weizhu Chen, Xiaodong Liu, Jianfeng
  Gao, and Jiawei Han.
\newblock On the variance of the adaptive learning rate and beyond, 2019.

\bibitem[Luenberger et~al.(1984)Luenberger, Ye, et~al.]{nonlinear_programming}
David~G Luenberger, Yinyu Ye, et~al.
\newblock \emph{Linear and nonlinear programming}, volume~2.
\newblock Springer, 1984.

\bibitem[Luo et~al.(2019)Luo, Xiong, and Liu]{AdaBound}
Liangchen Luo, Yuanhao Xiong, and Yan Liu.
\newblock Adaptive gradient methods with dynamic bound of learning rate.
\newblock In \emph{International Conference on Learning Representations}, 2019.
\newblock URL \url{https://openreview.net/forum?id=Bkg3g2R9FX}.

\bibitem[Mahsereci \& Hennig(2017)Mahsereci and
  Hennig]{probabilisticLineSearch}
Maren Mahsereci and Philipp Hennig.
\newblock Probabilistic line searches for stochastic optimization.
\newblock \emph{Journal of Machine Learning Research}, 18, 2017.

\bibitem[Martens \& Grosse(2015)Martens and Grosse]{KFAC}
James Martens and Roger Grosse.
\newblock Optimizing neural networks with kronecker-factored approximate
  curvature.
\newblock In \emph{International conference on machine learning}, pp.\
  2408--2417, 2015.

\bibitem[Mutschler \& Zell(2020)Mutschler and Zell]{pal}
Maximus Mutschler and Andreas Zell.
\newblock Parabolic approximation line search for dnns.
\newblock \emph{arXiv preprint arXiv:1903.11991}, 2020.

\bibitem[Ramamurthy \& Duffy(2017)Ramamurthy and Duffy]{L-sr1}
Vivek Ramamurthy and Nigel Duffy.
\newblock L-sr1: a second order optimization method.
\newblock 2017.

\bibitem[Reddi et~al.(2018)Reddi, Kale, and Kumar]{AmsGrad}
Sashank~J. Reddi, Satyen Kale, and Sanjiv Kumar.
\newblock On the convergence of adam and beyond.
\newblock In \emph{International Conference on Learning Representations}, 2018.
\newblock URL \url{https://openreview.net/forum?id=ryQu7f-RZ}.

\bibitem[Robbins \& Monro(1951)Robbins and Monro]{grad_descent}
H.~Robbins and S.~Monro.
\newblock A stochastic approximation method.
\newblock \emph{Annals of Mathematical Statistics}, 22:\penalty0 400--407,
  1951.

\bibitem[Rolinek \& Martius(2018)Rolinek and Martius]{L4}
Michal Rolinek and Georg Martius.
\newblock L4: Practical loss-based stepsize adaptation for deep learning.
\newblock In \emph{Advances in Neural Information Processing Systems}, pp.\
  6434--6444, 2018.

\bibitem[Schmidt et~al.(2020)Schmidt, Schneider, and
  Hennig]{schmidt2020descending}
Robin~M Schmidt, Frank Schneider, and Philipp Hennig.
\newblock Descending through a crowded valley--benchmarking deep learning
  optimizers.
\newblock \emph{arXiv preprint arXiv:2007.01547}, 2020.

\bibitem[Schraudolph et~al.(2007)Schraudolph, Yu, and Günter]{oLBFGS}
Nicol~N. Schraudolph, Jin Yu, and Simon Günter.
\newblock A stochastic quasi-newton method for online convex optimization.
\newblock In Marina Meila and Xiaotong Shen (eds.), \emph{Proceedings of the
  Eleventh International Conference on Artificial Intelligence and Statistics},
  volume~2 of \emph{Proceedings of Machine Learning Research}, pp.\  436--443,
  San Juan, Puerto Rico, 21--24 Mar 2007. PMLR.

\bibitem[Tieleman \& Hinton(2012)Tieleman and Hinton]{rmsProp}
Tijmen Tieleman and Geoffrey Hinton.
\newblock Lecture 6.5-rmsprop, coursera: Neural networks for machine learning.
\newblock \emph{University of Toronto, Technical Report}, 2012.

\bibitem[Vaswani et~al.(2019)Vaswani, Mishkin, Laradji, Schmidt, Gidel, and
  Lacoste-Julien]{backtracking_line_search_NIPS}
Sharan Vaswani, Aaron Mishkin, Issam Laradji, Mark Schmidt, Gauthier Gidel, and
  Simon Lacoste-Julien.
\newblock Painless stochastic gradient: Interpolation, line-search, and
  convergence rates.
\newblock \emph{arXiv preprint arXiv:1905.09997}, 2019.

\bibitem[Xing et~al.(2018)Xing, Arpit, Tsirigotis, and Bengio]{walkwithsgd}
Chen Xing, Devansh Arpit, Christos Tsirigotis, and Yoshua Bengio.
\newblock A walk with sgd.
\newblock \emph{arXiv preprint arXiv:1802.08770}, 2018.

\bibitem[Zeiler(2012)]{adadelta}
Matthew~D. Zeiler.
\newblock Adadelta: An adaptive learning rate method.
\newblock \emph{CoRR}, abs/1212.5701, 2012.

\end{thebibliography}
}
\vfill
\appendix
\pagebreak

\section{The full ELF algorithm}
\label{app:elf}
\begin{algorithm}[h!]
\caption{The ELF algorithm }
\begin{algorithmic}[1]
\label{algorithm:elf}
\renewcommand{\algorithmicrequire}{\textbf{Input:}}
\newcommand{\algorithmicbreak}{\textbf{break}}
\newcommand{\BREAK}{\STATE \algorithmicbreak}

\REQUIRE window\_size (default: 150)
\REQUIRE steps\_to\_train
\REQUIRE loss\_improvement\_faktor (default: 0.01)

\STATE losses $\leftarrow$  []
\STATE last\_mean\_loss $\leftarrow$ 0
\STATE t\_of\_last\_update $\leftarrow$ -1
\STATE expected\_per\_step\_improvement $\leftarrow$  $\infty$
\STATE t $\leftarrow$ 0
\STATE update\_step $\leftarrow$ 0

\WHILE{$t <$ steps\_to\_train}
	\STATE real\_improvement $\leftarrow$ last\_mean\_loss - mean(losses)
	\STATE expected\_improvement $\leftarrow$ last\_mean\_loss $-$ \\ expected\_per\_step\_improvement $\cdot$ (t$-$t\_of\_last\_update)
	\IF{($t$ $-$ t\_of\_last\_update$+$1) $\bmod (\text{windowsize}+1)==0$ and \\real\_improvement $\leq $ expected\_improvement $\cdot$ loss\_improvement\_faktor}
	\STATE suggested\_update\_step, expected\_per\_step\_improvement, \\ num\_batches\_loaded\_for\_line\_search $\leftarrow$ perform\_ELF\_line\_search() 
	\COMMENT{Algorithm \ref{algorithm:ELF_line_search}}
	\IF {suggested\_update\_step $> 0$}
	\STATE update\_step $\leftarrow$ suggested\_update\_step
	\ENDIF
	\STATE last\_mean\_loss $\leftarrow$ mean(losses)
	\STATE losses $\leftarrow$  []
	\STATE $t$ $\leftarrow$ $t$ $+$ num\_batches\_loaded\_for\_line\_search
	\STATE t\_of\_last\_update $\leftarrow t$
	%update t
	\ELSE
	\STATE loss $\leftarrow$ perform\_SGD\_training\_step(update\_step)
	\STATE losses.append(loss)
	\STATE $t$ $\leftarrow$ $t+1$ 
	\ENDIF
\ENDWHILE
\end{algorithmic} 
\end{algorithm}

\section{Further experimental details}
\label{app:hyperparam_emp_loss_comp}

For the experiments in Section \ref{subsec:emplosscomparison} the following non default hyperparameters were considered:\\
GOLS1: initial step size: 0.001, 0.01, 0.1, 1\\
PLS:  initial step size = 0.001, 0.01, 0.1, 1\\
ELF:  momentum = 0.0, 0.4 , decrease factor = 0, 0.2 \\

The training steps for the experiments in section Section \ref{subsec:robustness} were on all datasets except of ImageNet 100000 steps for DenseNet and 150000 steps for MobileNetv2, ResNet-18 and ResNet-34. On ImageNet 1687500 steps were trained.
The batch size was 128 for all experiments, except for ImageNet where it was 90.
The validation/train set splits were: 
5000/45000 for CIFAR-10 and CIFAR-100 
15000/60000 for Fashion MNIST
8257/65000 for SVHN
12000/1080000 for ImageNet.

All images were normalized with a mean and standard deviation determined over the dataset.
On CIFAR-10, CIFAR-100 and SVHN we used random horizontal flips and random cropping with size 32 and with a padding of 8 for CIFAR-100 and of 4 for SVHN and CIFAR-10.
On ImageNet we used random horizontal flips and and random resized crops with size 224.

To be compatible with the PLS implementation, Tensorflow 1.3 was used for the experiments in Section \ref{subsec:emplosscomparison}. Pytorch 1.6 was used for all other experiments.
\section{Training Loss and Validation accuracy plots of the experiments in Section \ref{subsec:robustness}}
\label{app:train_eval_plots}
\begin{figure}[h!]
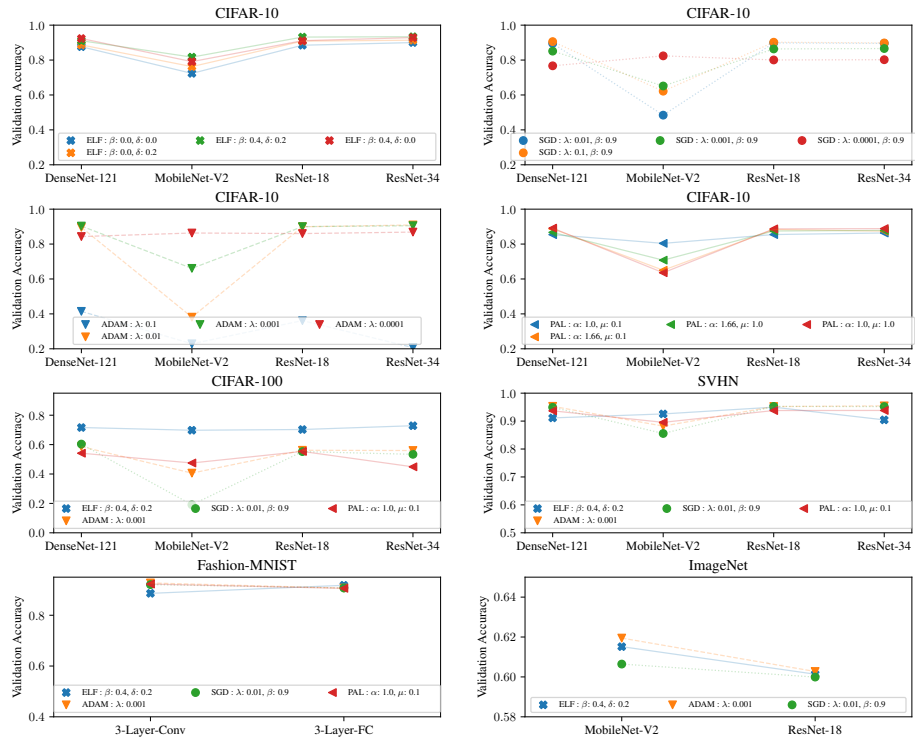

\centering
\scalebox{0.475}{
\input{plot_data/optimizer_comparison/plots/CIFAR-10_ELF_val.pgf}}\hspace{-0.5cm}
\scalebox{0.475}{
\input{plot_data/optimizer_comparison/plots/CIFAR-10_SGD_val.pgf}}\\
\scalebox{0.475}{
\input{plot_data/optimizer_comparison/plots/CIFAR-10_ADAM_val.pgf}}\hspace{-0.5cm}
\scalebox{0.475}{
\input{plot_data/optimizer_comparison/plots/CIFAR-10_PAL_val.pgf}}\\
\scalebox{0.475}{
\input{plot_data/optimizer_comparison/plots/CIFAR-100_val.pgf}}\hspace{-0.5cm}
\scalebox{0.475}{
\input{plot_data/optimizer_comparison/plots/SVHN_val.pgf}}
\scalebox{0.475}{
\input{plot_data/optimizer_comparison/plots/Fashion-MNIST_val.pgf}}\hspace{-0.5cm}
\scalebox{0.475}{
\input{plot_data/optimizer_comparison/plots/ImageNet_val.pgf}}

\caption{Corresponding Validation Accuracies to Figure \ref{fig:robustness_comparison}. ($\beta$=momentum, $\lambda$=learning rate, $\delta$=decrease factor) }
\label{fig:robustness_comparison_val}
\end{figure}
\vspace{-30cm}
\begin{figure}[t!]
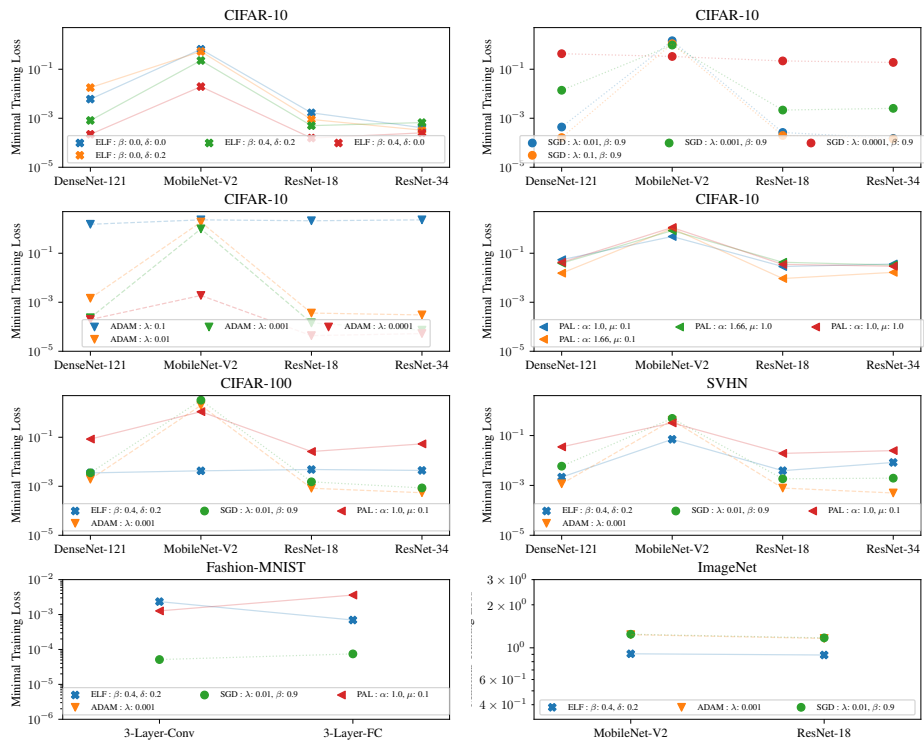

\centering
\scalebox{0.475}{
\input{plot_data/optimizer_comparison/plots/CIFAR-10_ELF_train.pgf}}\hspace{-0.5cm}
\scalebox{0.475}{
\input{plot_data/optimizer_comparison/plots/CIFAR-10_SGD_train.pgf}}\\
\scalebox{0.475}{
\input{plot_data/optimizer_comparison/plots/CIFAR-10_ADAM_train.pgf}}\hspace{-0.5cm}
\scalebox{0.475}{
\input{plot_data/optimizer_comparison/plots/CIFAR-10_PAL_train.pgf}}\\
\scalebox{0.475}{
\input{plot_data/optimizer_comparison/plots/CIFAR-100_train.pgf}}\hspace{-0.5cm}
\scalebox{0.475}{
\input{plot_data/optimizer_comparison/plots/SVHN_train.pgf}}
\scalebox{0.475}{
\input{plot_data/optimizer_comparison/plots/Fashion-MNIST_train.pgf}}\hspace{-0.5cm}
\scalebox{0.475}{
\input{plot_data/optimizer_comparison/plots/ImageNet_train.pgf}}

\caption{Corresponding training losses to Figure \ref{fig:robustness_comparison}. ($\beta$=momentum, $\lambda$=learning rate, $\delta$=decrease factor) }
\label{fig:robustness_comparison_train}
\end{figure}

\vfill \clearpage
\section{Further polynomial line approximations}
\label{app:furtherLineApp}
\begin{figure}[h!]
\includegraphics[width=0.24\linewidth]{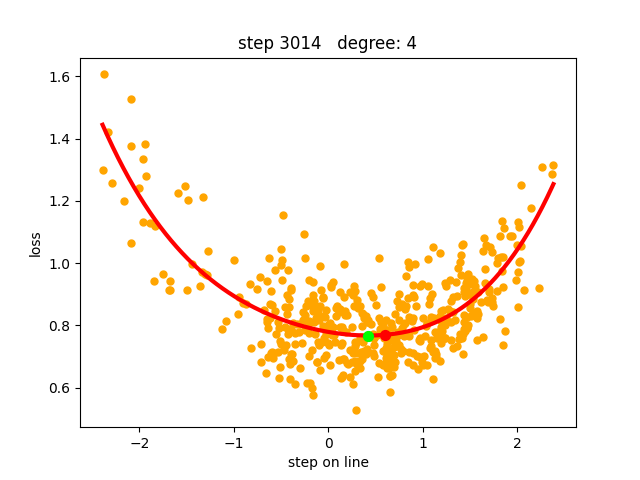}
\includegraphics[width=0.24\linewidth]{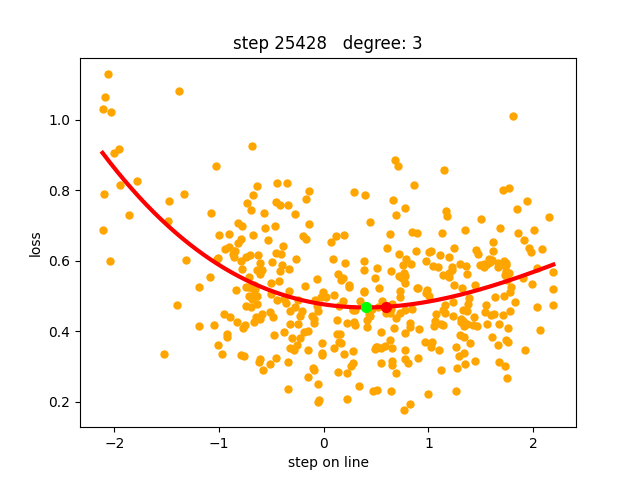}
\includegraphics[width=0.24\linewidth]{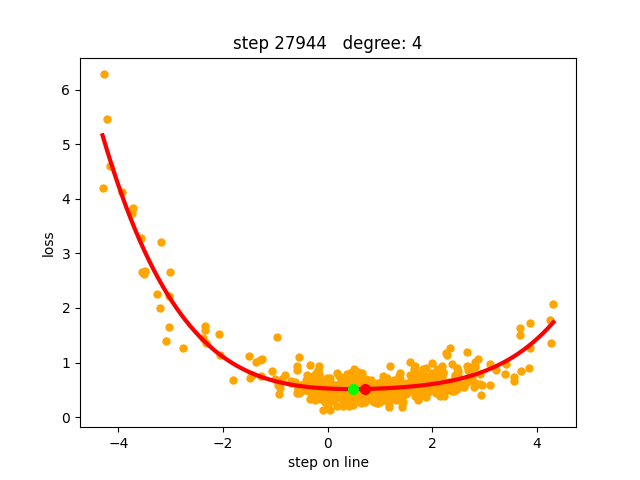}
\includegraphics[width=0.24\linewidth]{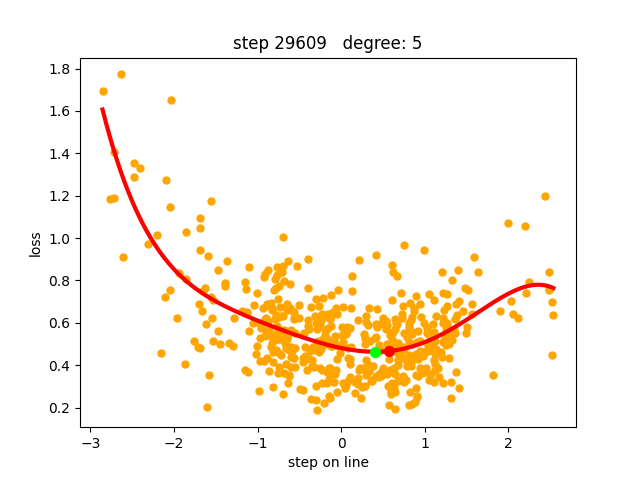}\\

\includegraphics[width=0.24\linewidth]{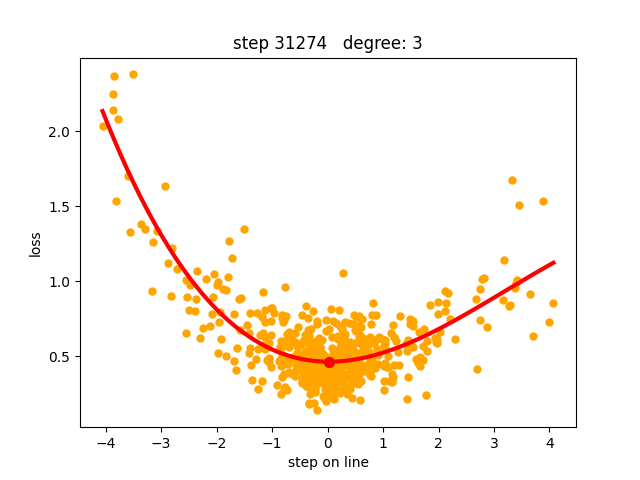}
\includegraphics[width=0.24\linewidth]{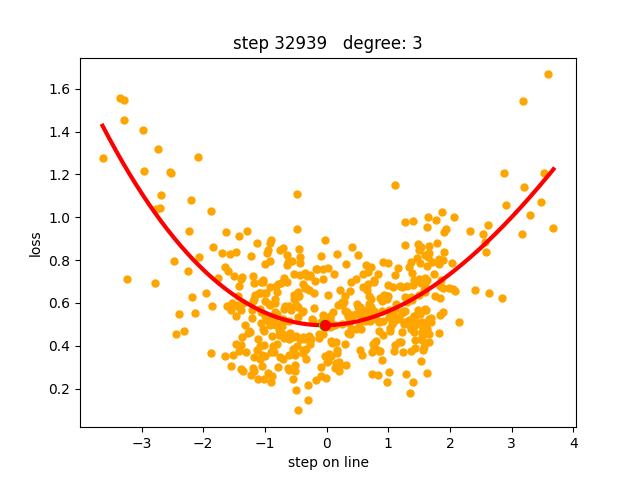}
\includegraphics[width=0.24\linewidth]{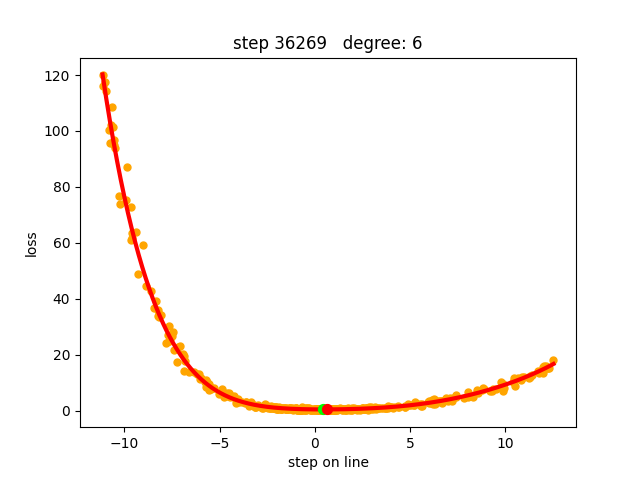}
\includegraphics[width=0.24\linewidth]{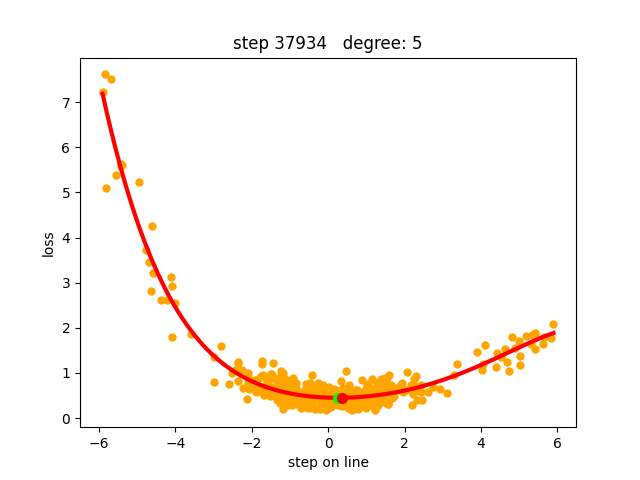}\\

\includegraphics[width=0.24\linewidth]{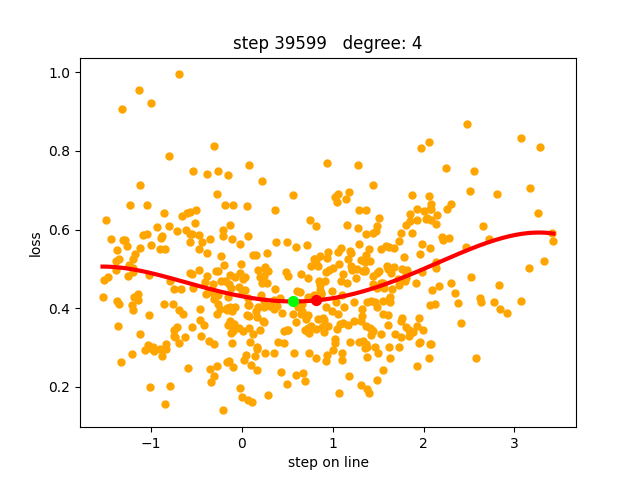}
\includegraphics[width=0.24\linewidth]{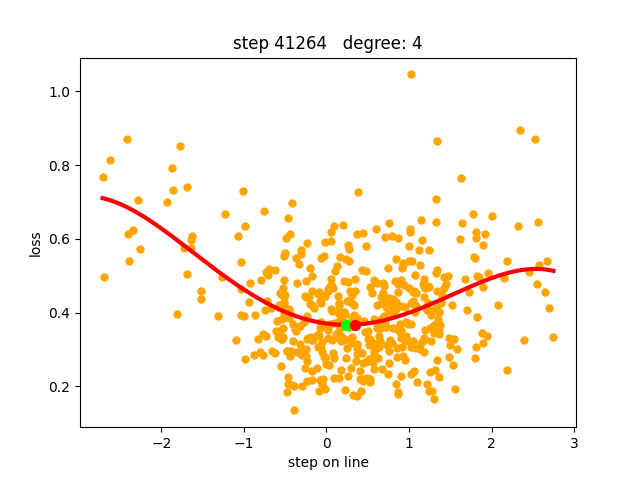}
\includegraphics[width=0.24\linewidth]{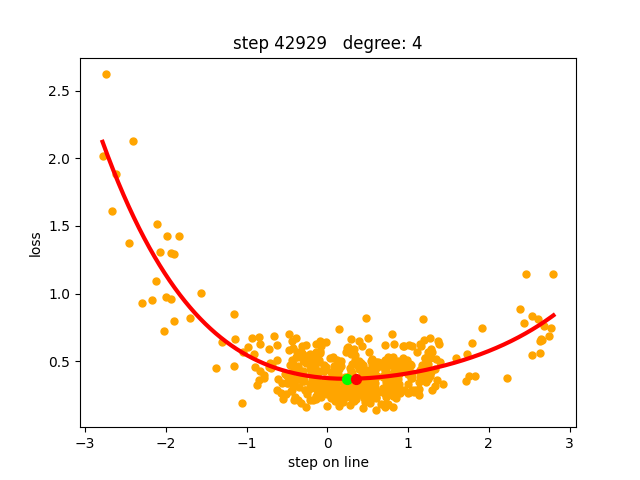}
\includegraphics[width=0.24\linewidth]{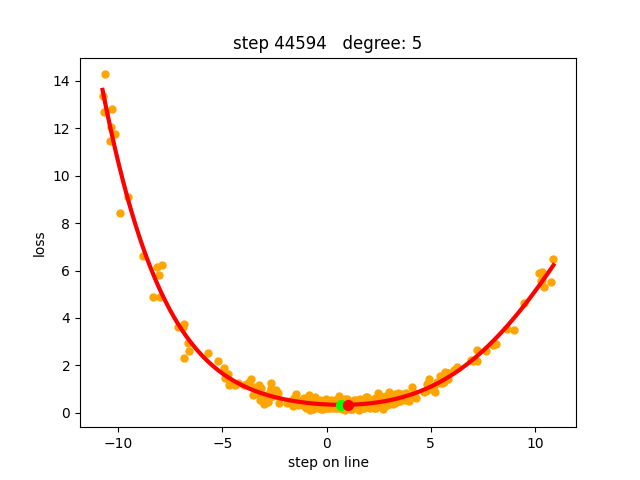}\\

\includegraphics[width=0.24\linewidth]{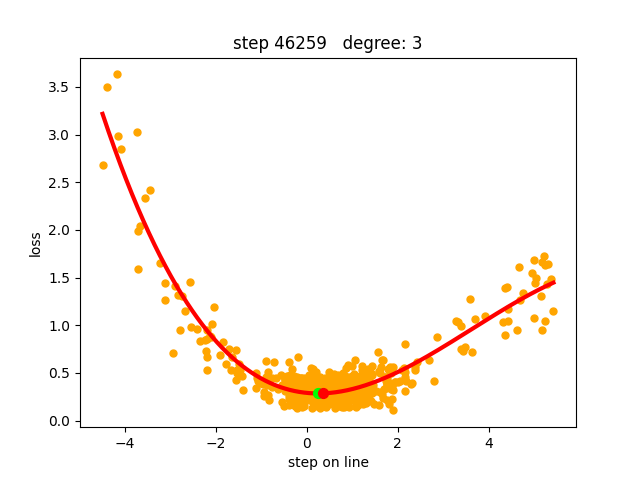}
\includegraphics[width=0.24\linewidth]{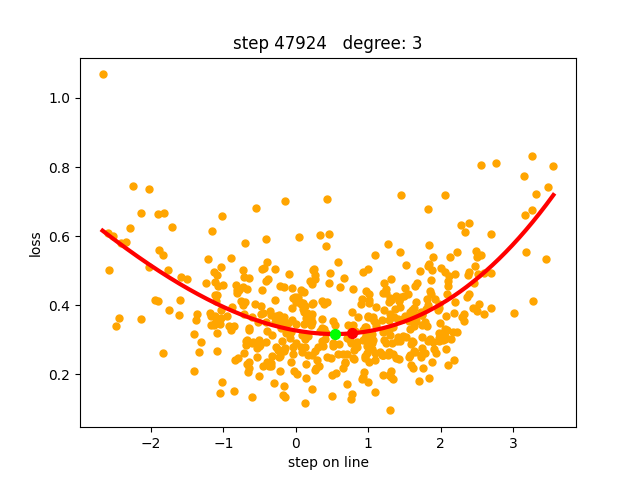}
\includegraphics[width=0.24\linewidth]{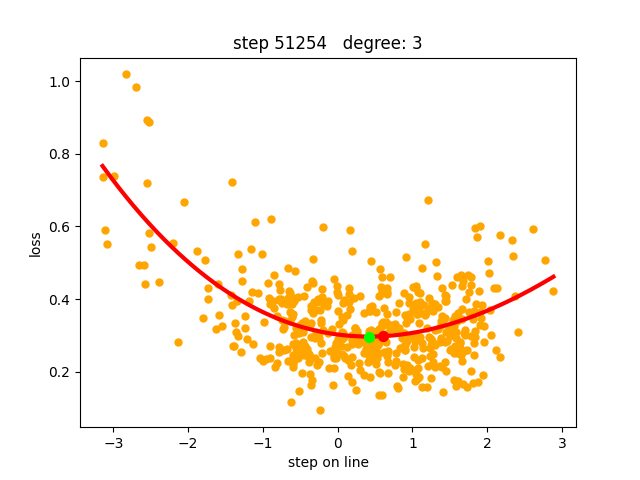}
\includegraphics[width=0.24\linewidth]{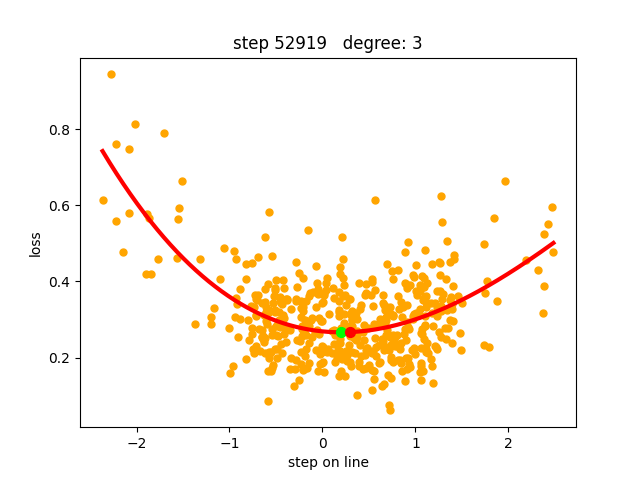}\\

\includegraphics[width=0.24\linewidth]{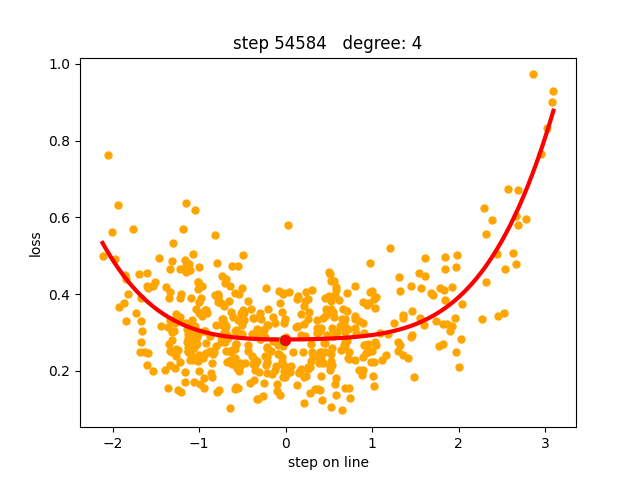}
\includegraphics[width=0.24\linewidth]{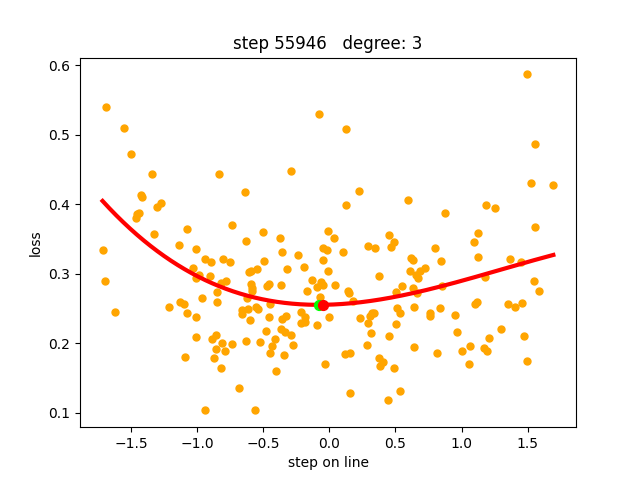}
\caption{Polynomial line approximations (\textcolor{red}{red}) from a training of a DenseNet on CIFAR-10. The samples losses are \textcolor{YellowOrange}{orange}. The minimum of the approximation is the \textcolor{green}{green dot}. The update step adjusted by a decrease factor of 0.2  is the  \textcolor{red}{red dot}.}
\end{figure}

\begin{figure}
\includegraphics[width=0.24\linewidth]{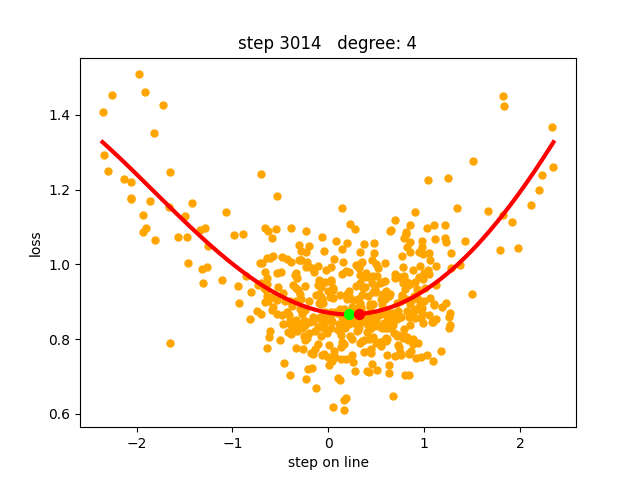}
\includegraphics[width=0.24\linewidth]{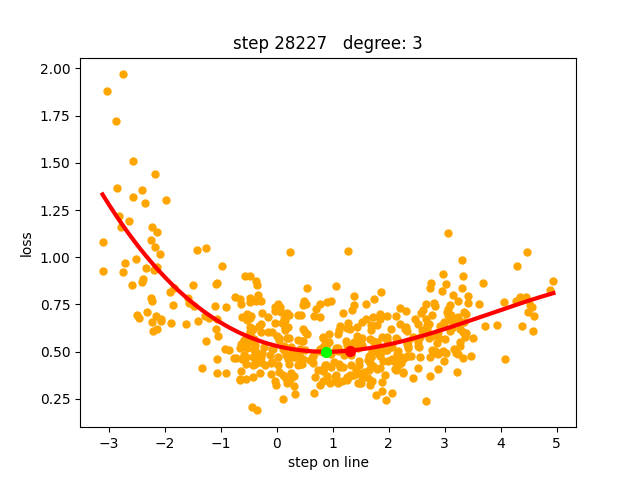}
\includegraphics[width=0.24\linewidth]{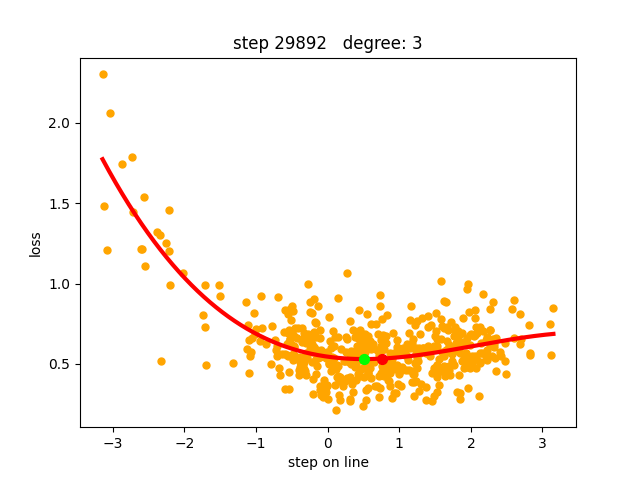}
\includegraphics[width=0.24\linewidth]{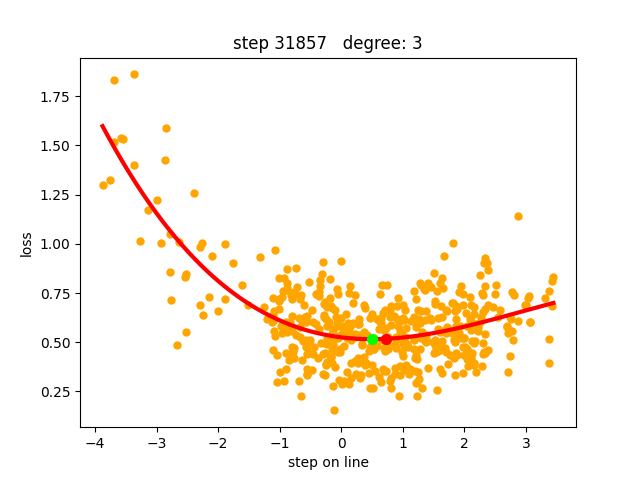}\\

\includegraphics[width=0.24\linewidth]{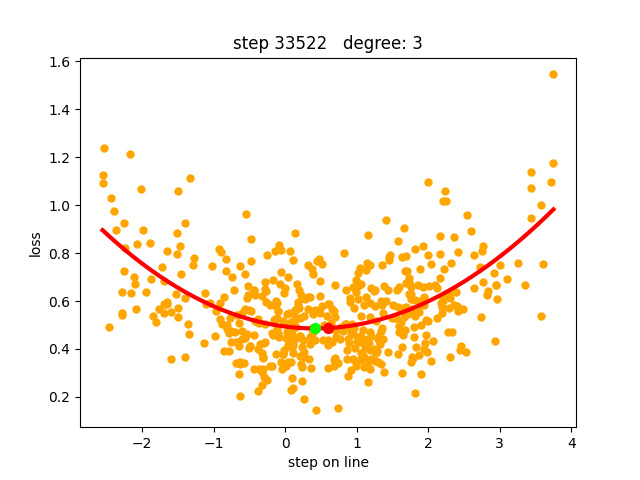}
\includegraphics[width=0.24\linewidth]{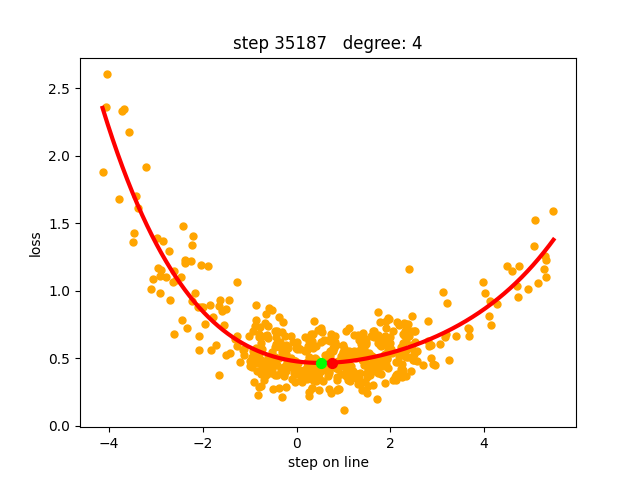}
\includegraphics[width=0.24\linewidth]{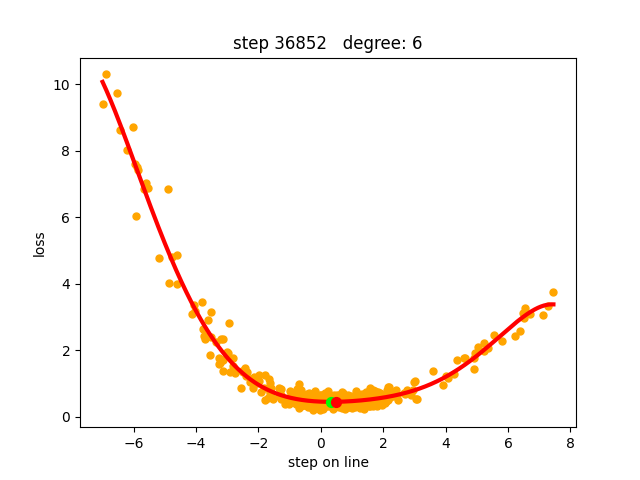}
\includegraphics[width=0.24\linewidth]{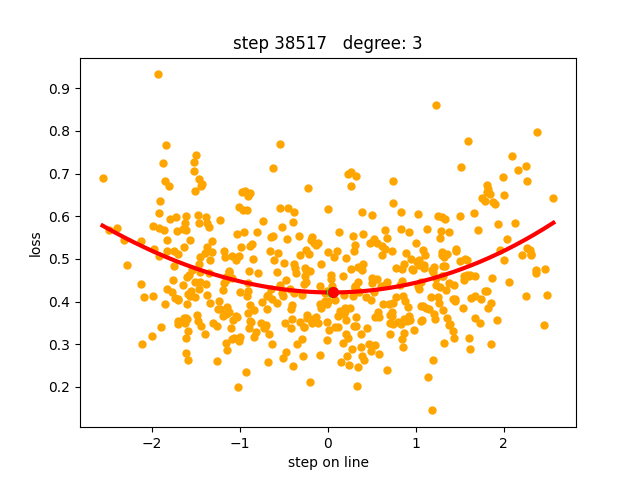}\\

\includegraphics[width=0.24\linewidth]{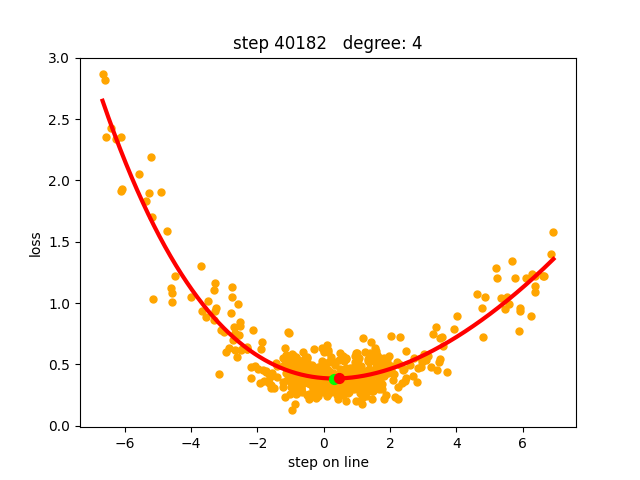}
\includegraphics[width=0.24\linewidth]{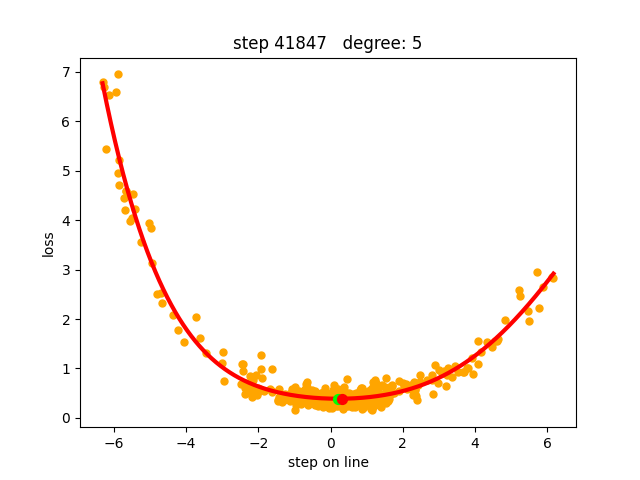}
\includegraphics[width=0.24\linewidth]{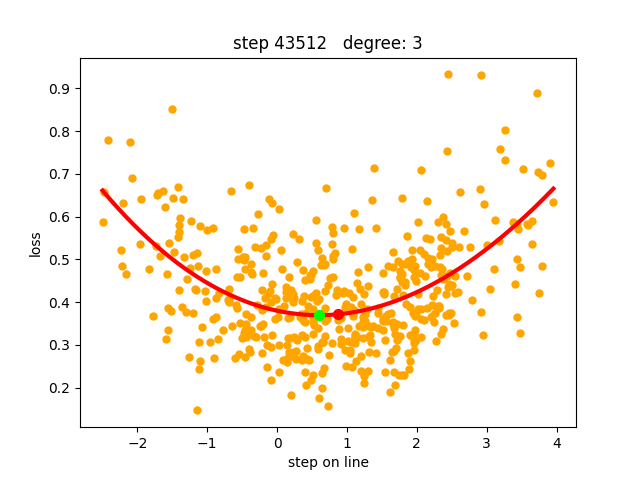}
\includegraphics[width=0.24\linewidth]{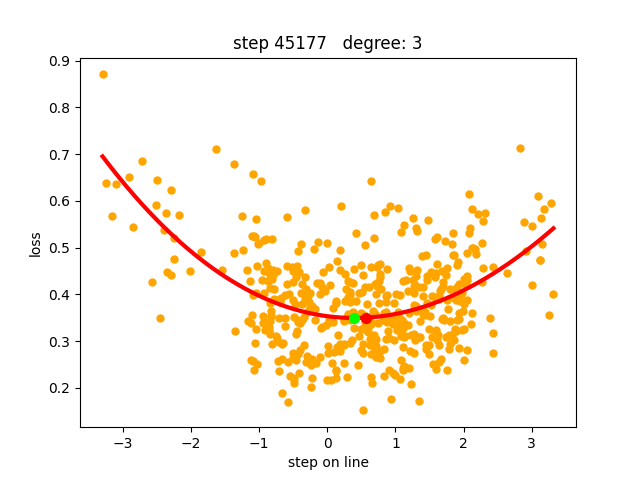}\\

\includegraphics[width=0.24\linewidth]{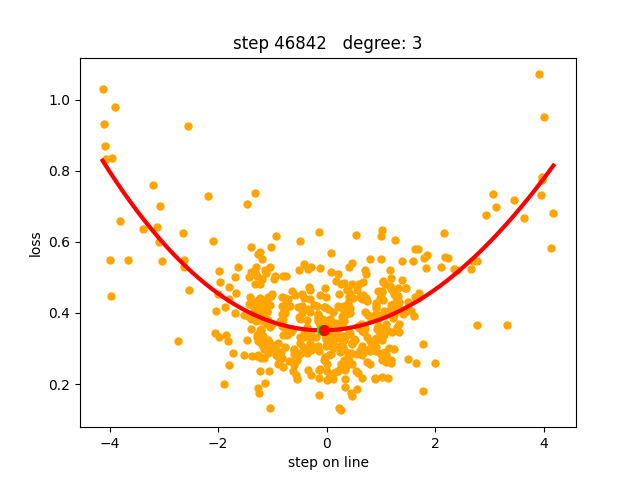}
\includegraphics[width=0.24\linewidth]{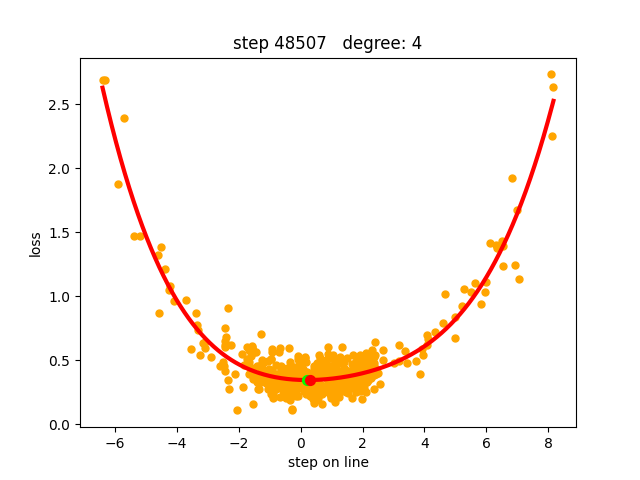}
\includegraphics[width=0.24\linewidth]{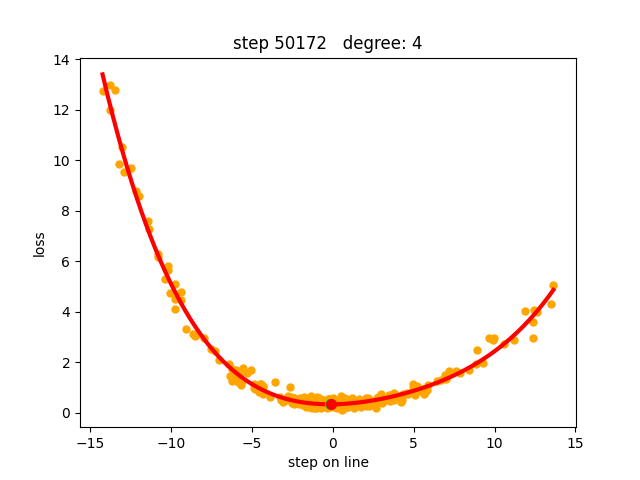}
\includegraphics[width=0.24\linewidth]{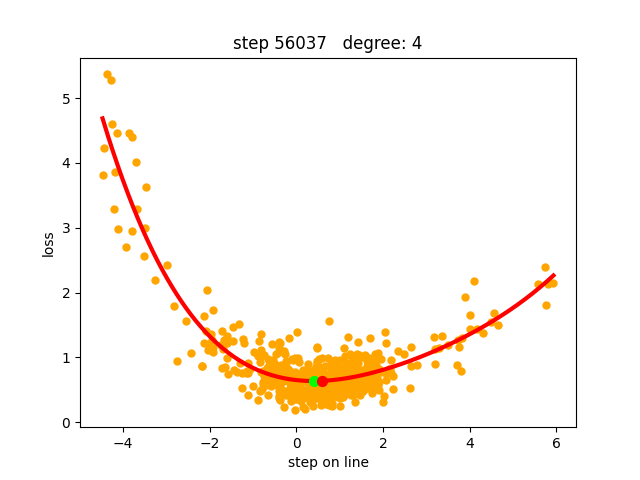}\\

\includegraphics[width=0.24\linewidth]{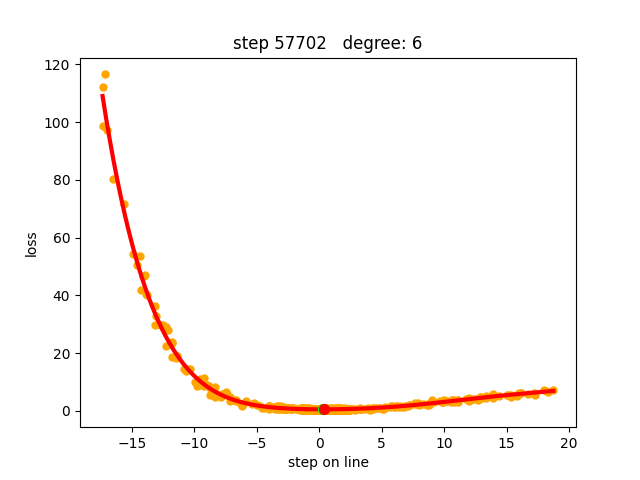}
\includegraphics[width=0.24\linewidth]{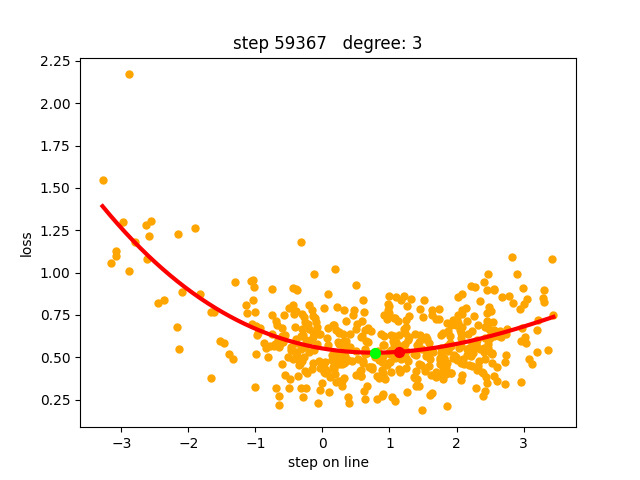}
\includegraphics[width=0.24\linewidth]{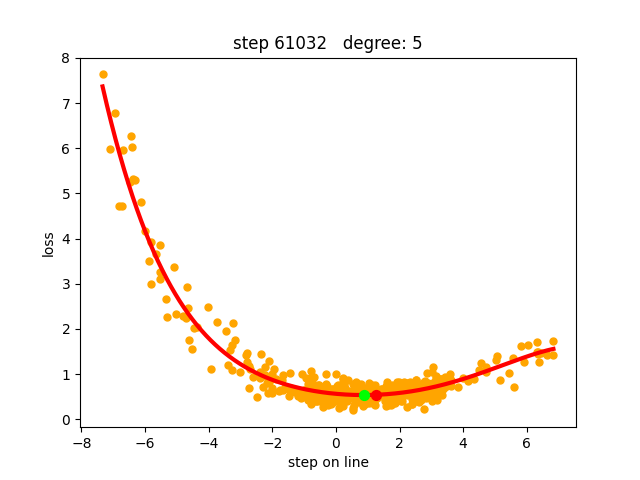}
\includegraphics[width=0.24\linewidth]{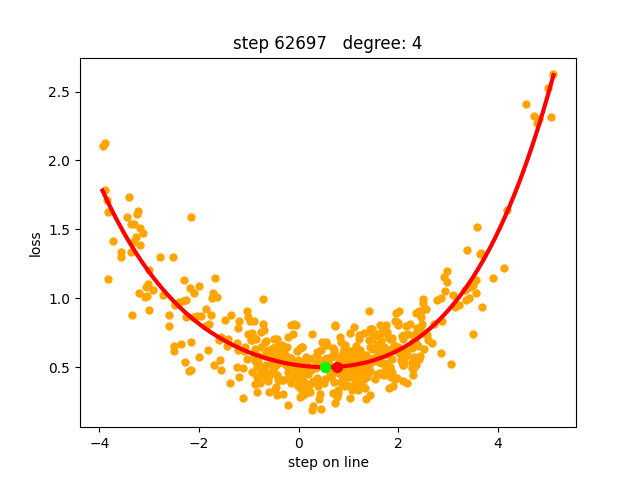}\\

\includegraphics[width=0.24\linewidth]{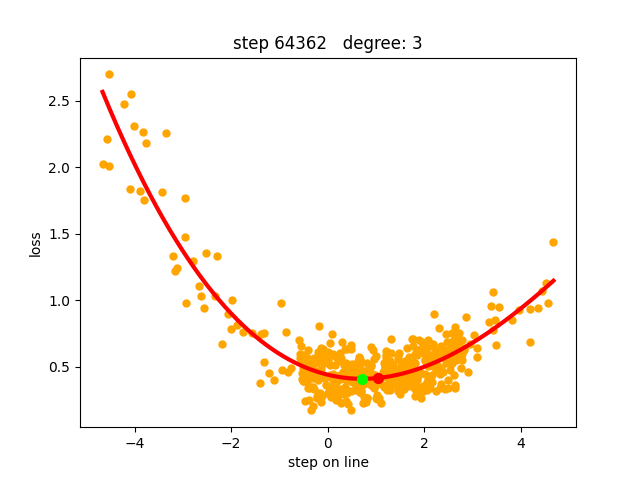}
\includegraphics[width=0.24\linewidth]{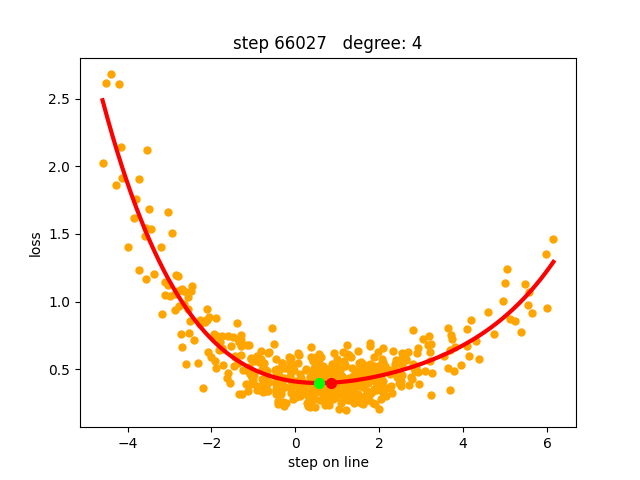}
\includegraphics[width=0.24\linewidth]{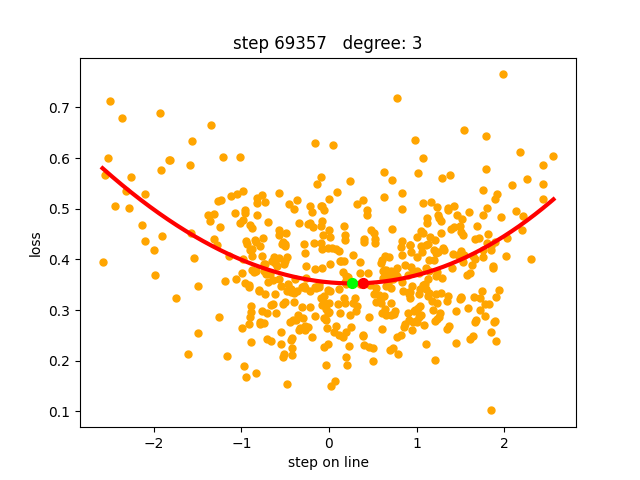}
\includegraphics[width=0.24\linewidth]{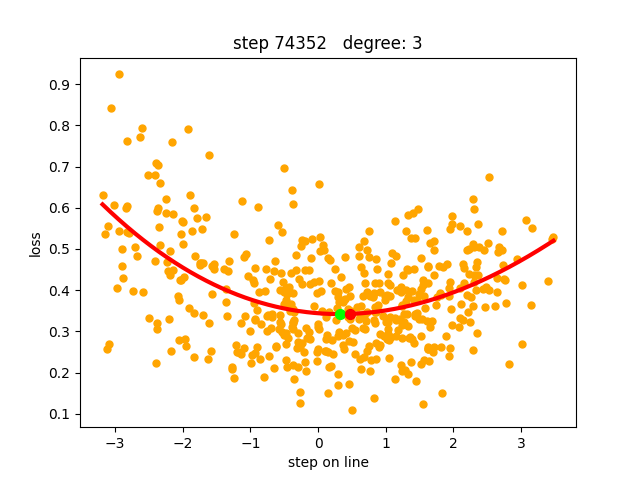}\\

\includegraphics[width=0.24\linewidth]{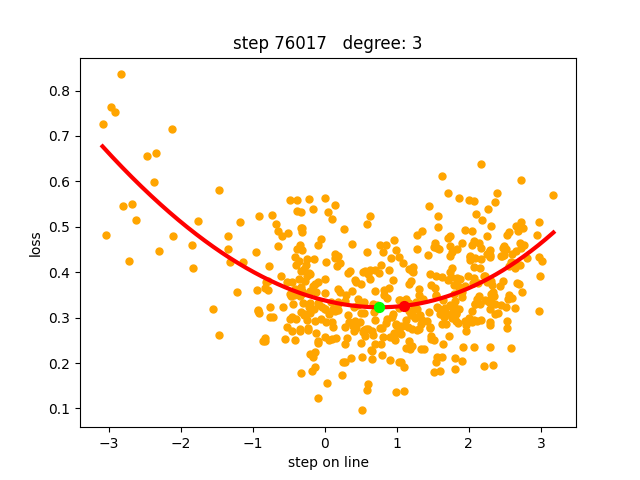}
\includegraphics[width=0.24\linewidth]{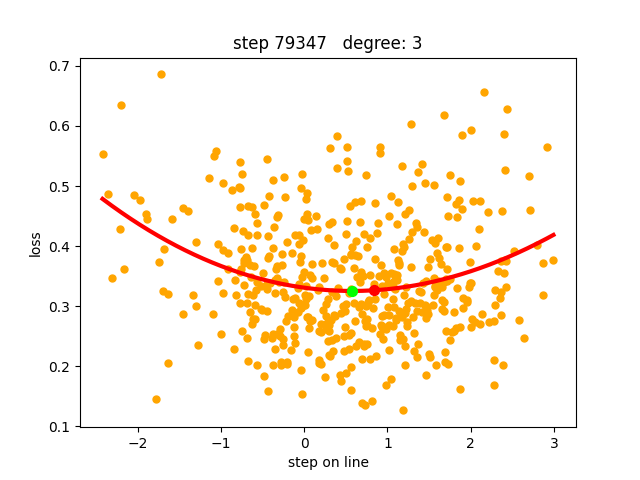}
\includegraphics[width=0.24\linewidth]{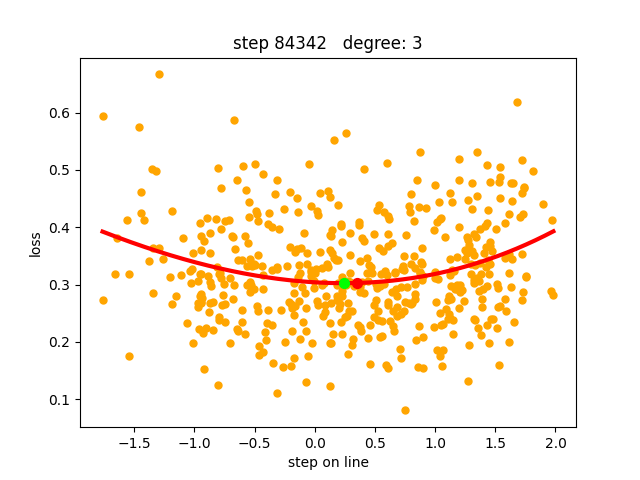}
\includegraphics[width=0.24\linewidth]{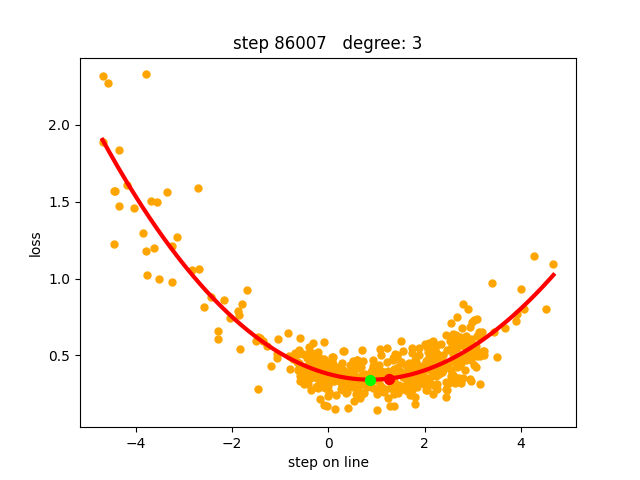}\\

\caption{Polynomial line approximations (\textcolor{red}{red}) from a training of a ResNet18 on CIFAR-10. The samples losses are \textcolor{BurntOrange}{orange}. The minimum of the approximation is the \textcolor{green}{green dot}. The update step adjusted by a decrease factor of 0.2  is the  \textcolor{red}{red dot}.}
\end{figure}

\end{document}